\pdfoutput=1

\documentclass[11pt]{article}

\usepackage[]{acl}

\usepackage{times}
\usepackage{latexsym}
\usepackage{xcolor}

\usepackage[T1]{fontenc}

\usepackage[utf8]{inputenc}
\usepackage{url}

\usepackage{microtype}
\usepackage[most]{tcolorbox}
\usepackage{cuted}
\tcbuselibrary{breakable}
\usepackage{booktabs}
\usepackage{multirow}
\usepackage{graphicx}
\usepackage{tabularx}
\usepackage{subcaption}
\usepackage{longtable}
\usepackage{hyperref}
\usepackage{threeparttable}
\usepackage[ruled, lined, linesnumbered, commentsnumbered, longend]{algorithm2e}

\usepackage{amsmath, amsmath}
\usepackage{makecell}
\usepackage{multirow}
\usepackage{array}
\usepackage{soul}
\usepackage[normalem]{ulem}
\usepackage{listings}
\usepackage[switch]{lineno}  %
\usepackage{enumitem}
\usepackage{amssymb}
\usepackage{bbm}

\title{\texttt{DARA}: Decomposition-Alignment-Reasoning Autonomous Language Agent for Question Answering over Knowledge Graphs}

\author{Haishuo Fang\textsuperscript{1} \quad Xiaodan Zhu\textsuperscript{1,2} \quad Iryna Gurevych\textsuperscript{1}\\ 
  \textsuperscript{1}Ubiquitous Knowledge Processing Lab (UKP Lab), Department of Computer Science and \\
  Hessian Center for AI (hessian.AI), Technical University of Darmstadt, Germany \\
  \textsuperscript{2}Department of Electrical and Computer Engineering \& Ingenuity Labs Research Institute,  \\
  Queen’s University, Canada \\
  \textsuperscript{1}\texttt{\href{www.ukp.tu-darmstadt.de}{www.ukp.tu-darmstadt.de}} \quad \textsuperscript{2}\texttt{\href{mailto:xiaodan.zhu@queensu.ca}{xiaodan.zhu@queensu.ca}}}
\begin{document}
\maketitle
\begin{abstract}
Answering Questions over Knowledge Graphs (KGQA) is key to well-functioning autonomous language agents in various real-life applications.
To improve the neural-symbolic reasoning capabilities of language agents powered by Large Language Models (LLMs) in KGQA, we propose the \textbf{D}ecomposition-\textbf{A}lignment-\textbf{R}easoning \textbf{A}gent (\texttt{DARA}) framework. \texttt{DARA} effectively parses questions into formal queries through a dual mechanism: high-level iterative task decomposition and low-level task grounding. Importantly, \texttt{DARA} can be efficiently trained with a small number of high-quality reasoning trajectories.
Our experimental results demonstrate that \texttt{DARA} fine-tuned on LLMs (e.g. Llama-2-7B, Mistral) outperforms both in-context learning-based agents with GPT-4 and alternative fine-tuned agents, across different benchmarks in zero-shot evaluation.
This makes such models more accessible for real-life applications.
We also show that \texttt{DARA} attains performance comparable to state-of-the-art enumerating-and-ranking-based methods for KGQA\footnote{Code, model are released at \url{https://github.com/UKPLab/acl2024-DARA}}.

\end{abstract}

\section{Introduction}\label{sec:intro}
\begin{figure}
    \centering
    \includegraphics[width=\columnwidth]{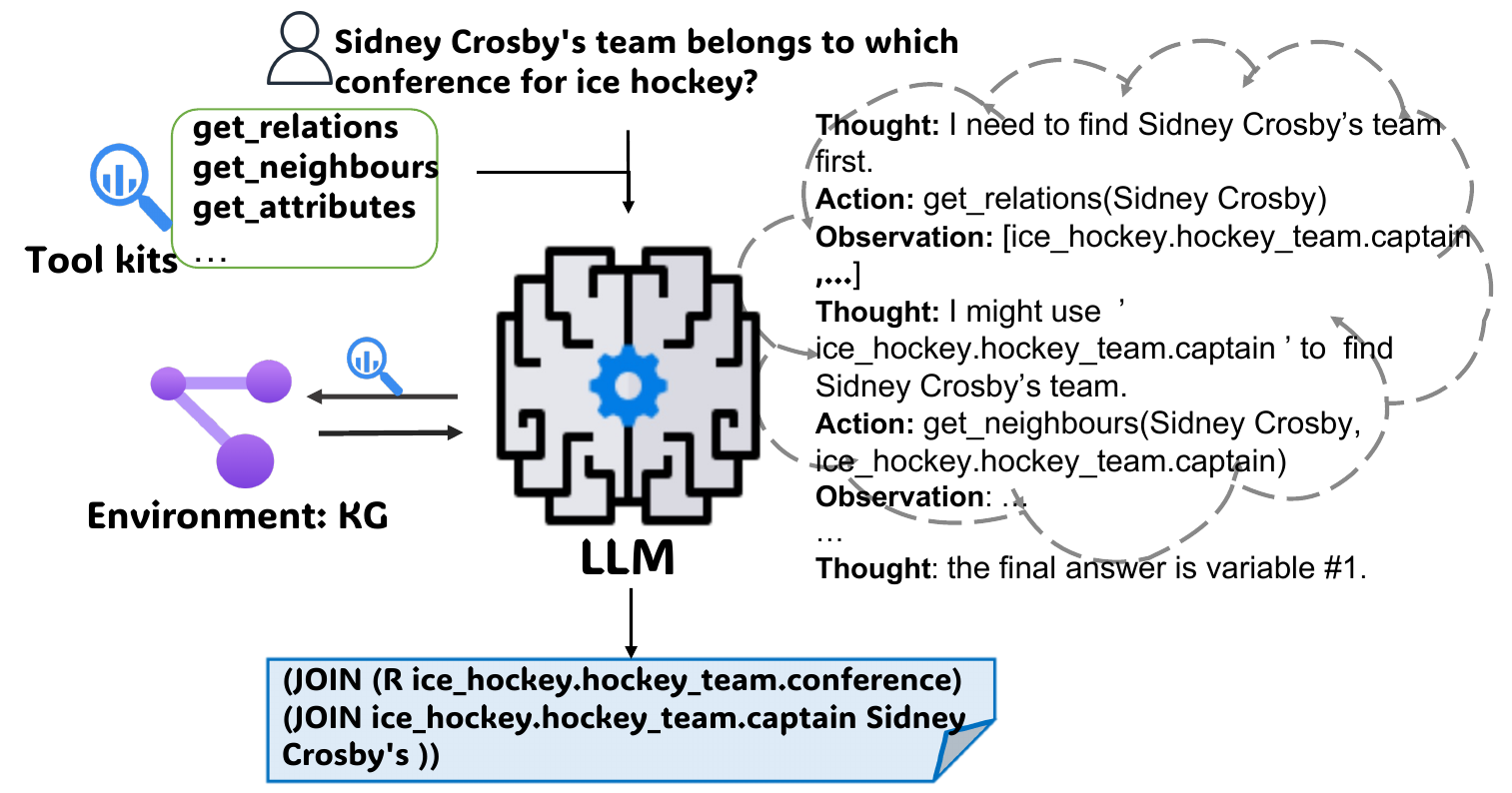}
    \caption{The pipeline of the language agent to parse the user's question into a logical form via proactive KG interaction and reasoning. 
    }
    \label{fig:language agent}
\end{figure}

Language agents~\cite{wang2023agent,theodore2023cola,xi2023agentsurvey} building on Large Language Models (LLMs) have emerged as a new
paradigm of artificial intelligence (AI) systems to perform complex tasks in various environments.
These agents primarily rely on LLMs as central controllers to perceive environments, autonomously plan, and reason to fulfill user tasks.
However, operating within environments rich in structured data, such as Knowledge Graphs (KGs)~\cite{kurt2008freebase, vrandecic2014wikidata},
presents unique challenges for the agents.
For instance, performing Question Answering over Knowledge Graphs (KGQA) demands various capabilities from language agents.
The agents need to decompose the user's question into subtasks (\textit{planning}), interact with the KG to obtain schemas (\textit{tool usage}), select appropriate schema items aligned with subtasks (\textit{schema alignment}), and construct executable logical forms deriving the answer (\textit{logical reasoning}) (Figure~\ref{fig:language agent}).
Previous research, e.g., tool learning with foundation models~\cite{qin2023tool} and AgentBench~\cite{liu2023agentbench}, has equipped off-the-shelf LLMs with multiple functions to conduct KGQA in the In-Context Learning (ICL) setup~\cite{tom2020gpt3}.
Although ICL-based agents are easy to use and finetuning-free, the performance of open-sourced ICL agents falls \textit{short significantly} compared to classical enumerating-and-ranking-based methods~\cite{shu-etal-2022-tiara,gu-etal-2023-dont}, in which all possible reasoning paths starting from anchor entities in the question are enumerated and ranked.
Despite this, LLMs-based agents continue to advance to be more powerful as LLMs are scaled up and rapidly improve every year.

While ICL-based language agents with GPT-4~\cite{openai2023gpt} show strong capabilities in this task, the use of commercial LLMs raises practical concerns related to privacy, costs, and model flexibility. For example, running AgentBench with GPT-4 over only \textasciitilde4,500 test examples incurs an expenditure of 1,300 U.S. dollars in stark contrast to \textasciitilde\$30 of the proposed model trained on open-sourced LLMs (see section~\ref{sec:expense} for details), which raises critical concerns for a variety of applications.

To enhance the agent capabilities of LLMs and circumvent the challenges of ICL-based LLM agents,
AgentTuning~\cite{zeng2023agenttuning} fine-tunes AgentLMs based on Llama-2~\cite{hugo2023llama2} using the AgentBench framework for six different environments including Knowledge Graphs.
However, the performance of fine-tuned AgentLMs still lags behind GPT-4 agents on structured data processing tasks including Knowledge Graphs, Databases, and Operating Systems.

In this paper, we propose the \textbf{D}ecomposition-\textbf{A}lignment-\textbf{R}easoning \textbf{A}gent (\texttt{DARA}) framework for the KGQA task.
\texttt{DARA} is a hierarchical framework, which comprises two connected modules: a high-level \textit{task decomposition} and low-level \textit{task grounding} module. The former is responsible for decomposing the given question into smaller tasks, while the latter derives executable logical forms for the decomposed task.
Within the task grounding module, two integral components contribute to the overall functionality. The \textit{schema item selection component} identifies appropriate schema items (relations, classes) relevant to the task, while the \textit{logical form construction component} assembles a logical form based on the selected schema as well as the preceding logical form.
To enhance the selection of relations, we propose a relation selection method, termed skim-then-deep-reading, where \texttt{DARA} scans relations of current entities and selects $n$ promising relations to deeply read their descriptions.
Task decomposition and task grounding work in tandem until the full logical form is 
constructed.

To generate reasoning trajectories for fine-tuning, we initially employ GPT-4 to translate linearized logical forms into natural language.
Unlike previous work~\cite{gilardi2023chatgpt, xu2023wizardlm} showed the success of GPT-4 in annotation, we found the conversion quality falls short of expectations and can be subsequently enhanced through human verification.
To assess the efficacy of \texttt{DARA}, we conduct comprehensive experiments on three popular benchmark datasets, i.e., WebQSP~\cite{yih-etal-2016-value}, GraphQ~\cite{su-etal-2016-generating}, and GrailQA~\cite{yu2021grailqa}.
Our findings confirm that \texttt{DARA} substantially outperforms both the ICL-based and the alternative fine-tuned LLM agents.
To conclude, our contributions are:

\begin{itemize}[leftmargin=5mm,labelsep=3mm]
    \item We propose a novel language agent framework for KGQA, \textbf{D}ecomposition-\textbf{A}lignment-\textbf{R}easoning \textbf{A}gent (\texttt{DARA}).
   It surpasses the framework proposed in AgentBench, by explicitly disentangling high-level task decomposition and low-level task grounding (\textit{schema items selection} and \textit{logical form construction}).
    \item Experiments show that fine-tuned \texttt{DARA} achieves state-of-the-art performance compared with both ICL-based and other fine-tuned agents (AgentLMs and fine-tuned AgentBench) across the three important benchmarks in zero-shot evaluation. Moreover, training with 768 reasoning trajectories, we show that \texttt{DARA} can achieve highly competitive performances comparable to enumerating-and-ranking-based models trained on larger data.
    
    \item Our experiments reveal the ongoing challenge of generating high-quality reasoning trajectories for language agents in KGQA with GPT-4.
    This is in contrast to previous studies that demonstrate the success of ChatGPT or GPT-4 in annotation~\cite{gilardi2023chatgpt, xu2023wizardlm}. This observation suggests a potential avenue for future research: how to automatically generate high-quality data for language agent use cases where the most advanced LLMs (e.g. GPT-4) face their limitations.
    
\end{itemize}
\section{Related work}\label{sec:related_work}
\textbf{KG-enhanced LLM Reasoning} is a popular paradigm to reduce hallucination and unfaithful reasoning chains of LLMs. In this approach, retrieved triplets from KGs and parametric knowledge within LLMs (i.e., knowledge stored in their parameters)~\cite{petroni-etal-2019-language,roberts-etal-2020-much} work in tandem to derive the final answer to a given question.
The Knowledge-Driven Chain-of-Thought framework~\cite{wang2023knowledge} refines LLM reasoning using an external QA model based on KGs.
Think-on-Graph~\cite{sun2023think} and StructGPT~\cite{jiang2023structgpt} utilize off-the-shelf LLMs to traverse over graphs to find the most relevant knowledge and integrate them with parametric knowledge of LLMs to produce the final answer.
Despite their potential, these methods face challenges when the parametric knowledge in LLMs is incorrect or outdated.
In addition, in scenarios of conflict between parametric knowledge and external non-parametric knowledge, ~\citet{qian2023merge} reveals that LLMs are prone to the distraction of external knowledge, when the latter is irrelevant, leading to worse performance.
\paragraph{LLM-based Autonomous Agents for KGQA.}
Unlike KG-enhanced LLM reasoning which focuses on refining reasoning chains using KGs, LLM agents are able to conduct more complex tasks such as constructing logical forms for a given question by utilizing a set of more powerful human-like capabilities~\cite{theodore2023cola} including question decomposition and logical reasoning.
Tool learning with foundation models~\cite{qin2023tool},  AgentBench~\cite{liu2023agentbench}, and AgentTuning~\cite{zeng2023agenttuning} are three representative works.
~\citet{qin2023tool} asked LLM agents to directly write SPARQL (Standard Protocol and RDF Query Language) program. In contrast, AgentBench adopts intermediate representation, namely \textit{s-expression}~\cite{yu2021grailqa}, to represent SPARQL and construct the \textit{s-expression} program step-by-step.
AgentTuning fine-tunes LLMs with reasoning trajectories. 
However, all of them achieve poor performance due to the limitations of ICL-based agents 
 and the framework design.

\paragraph{Enumerating-and-ranking-based Methods} are a prevalent paradigm for KGQA~\cite{yih-etal-2015-semantic,lan-jiang-2020-query,luo-etal-2018-knowledge,abujabal2023automated}. To narrow down the large search space of KGs, it first enumerates all possible candidate logical forms from neighborhood of topic entities or retrieve similar logic forms from training examples. 
Subsequently, a ranker is applied to select the best one.
However, this brute-force approach can lead to exponential candidates and thus suffer from scalability and coverage issues.
To alleviate these issues, generation-augmented methods~\cite{ye-etal-2022-rng,shu-etal-2022-tiara} generate diverse logical forms based on retrieved candidates. Besides, dynamic bottom-up semantic parsing approaches \cite{gu-etal-2023-dont,gu-su-2022-arcaneqa} construct the final logical form incrementally and prune the search space on the fly. 
Although these methods can achieve high performance, they are time-consuming and necessitate expert-crafted rules for logical form construction. 
In contrast, LLM-based agents leverage LLMs for planning, grounding, and reasoning, offering enhanced explainability, efficiency, and flexibility. Moreover, LLM agents automate the entire process from task decomposition to relation selection, to logical form construction, eliminating the need for extensive enumeration based on expert-crafted rules.


\section{The Approach}
\subsection{Overview}
\begin{figure*}
    \centering
    \includegraphics[width=0.9\textwidth]{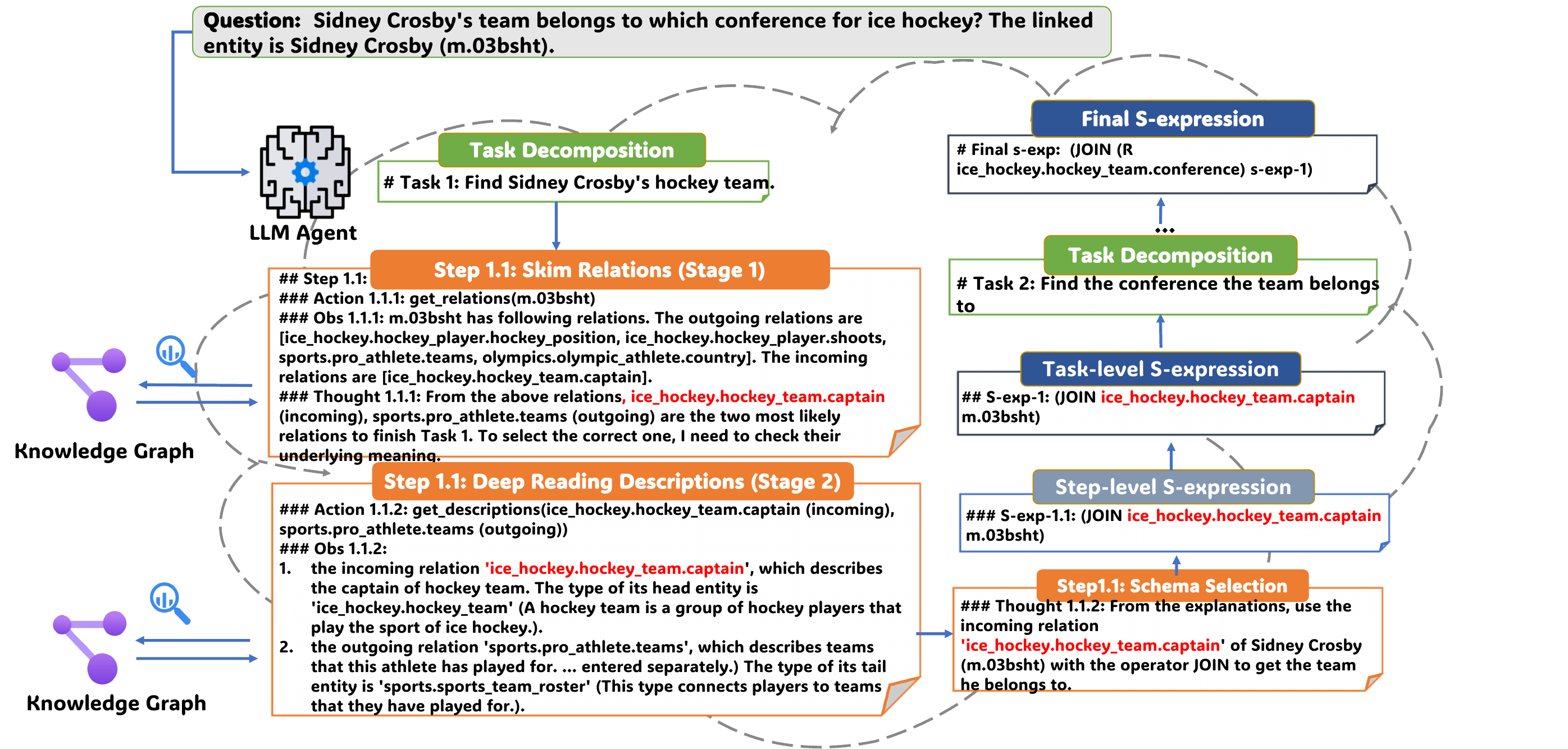}
    \caption{An example of \texttt{DARA} to automatically construct a logical form for the question \textit{Sidney Crosby's team belongs to which conference for ice hockey?} \textcolor{red}{$ice\_hockey.hockey\_team.captain$} is the selected relation using skim-then-deep-reading selection method.}
    \label{fig:overview} 
\end{figure*}

An overview of \texttt{DARA} is described in Algorithm~\ref{algo:dara}. Formally, given a knowledge graph $\mathcal{G}$, a question $\mathcal{Q}$, and a set of actions $\mathcal{A}$, the objective is to construct a logical form  $\mathcal{L}$, e.g., \textit{s-expression}~\cite{yu2021grailqa}, that yields the final answer to the question $\mathcal{Q}$.
To achieve this objective, \texttt{DARA} iteratively performs \textit{task decomposition} (Section~\ref{sec:ITD}) and \textit{task grounding} (Section~\ref{sec:task_grounding}) until $\mathcal{L}$ is finalized.
During the iteration $i$, for decomposed task $\mathcal{T}_{i}$ (\hyperref[line:td]{\textit{line 7}}), \texttt{DARA} grounds it against $\mathcal{G}$ (\hyperref[line:task_grounding]{\textit{line 16}}).
It takes actions in $\mathcal{A}$ to fetch schema items from $\mathcal{G}$ and selects the suitable one to construct the step-level logical form $\mathcal{L}_{ij}$ (Section~\ref{sec:task_grounding}).
This process iterates multiple times until $\mathcal{L}_{ij}$ can fulfill the task $\mathcal{T}_{i}$ (\hyperref[line:reasoning]{\textit{line 15-24}}).
Subsequently, $\mathcal{L}_{ij}$ will be assigned to $\mathcal{L}_{i}$ with the task level id (e.g. s-exp-1) (\hyperref[line:task-level-lf]{\textit{line 19}}). 
\texttt{DARA} incrementally construct the full logical form $\mathcal{L}$ via assembling task-level logical forms.
The base conditions for exiting loops are autonomously determined by \texttt{DARA}. A concrete example illustrating this process is presented in Figure~\ref{fig:overview}.

\begin{algorithm}[htb]
    \SetKwFunction{isOddNumber}{isOddNumber}
    \small
    \KwIn{A knowledge graph $\mathcal{G}$, a question $\mathcal{Q}$, a set of actions $\mathcal{A}$}
    \KwOut{Grounded logical form $\mathcal{L}$}
    \tcp{Reasoning Trajectory}
    $Traces \leftarrow \mathcal{Q}$ \;
    \tcp{Initial task-level logical form}
    $\mathcal{L}_{0} \leftarrow null$ \;
    $i \leftarrow 1$  \;
    \While{True}{
        $\mathcal{T}_{i} \leftarrow 
        \textbf{task\_decomposition}(Traces)$\; \label{line:td}
        \tcp{No further task needed}
        \If{$\mathcal{T}_{i}$ is null} {
            $\mathcal{L} \leftarrow \mathcal{L}_{i-1}$ \;
            break \;
        }{\label{line:task finish}}
        $Traces \leftarrow Traces + \mathcal{T}_{i}$\;
        $j \leftarrow 1$ \;
        \While{True}{ \label{line:reasoning}
            $\mathcal{L}_{ij} \leftarrow \textbf{task\_grounding}(Traces, \mathcal{A}, \mathcal{G})$ \;\label{line:task_grounding}
            $Traces \leftarrow Traces + \mathcal{L}_{ij}$\;
            \If{$\mathcal{L}_{ij}$ completes $\mathcal{T}_{i}$}{ \label{line:base_cond_grounding}
                $\mathcal{L}_{i} \leftarrow \textbf{replace\_id}(\mathcal{L}_{ij}$) \;\label{line:task-level-lf}
                $Traces \leftarrow Traces + \mathcal{L}_{j}$\

                break \;
            }
            $j \leftarrow j + 1$ \;
        }
        $i \leftarrow i + 1$ \;  
    }
    \Return $\mathcal{L}$
    \caption{\texttt{DARA}}
    \label{algo:dara}
\end{algorithm}

\subsection{The Reasoning Framework}
\subsubsection{Iterative Task Decomposition}\label{sec:ITD}
Iterative task decomposition serves as the high-level planner in our framework. Decomposing questions into executable tasks is a fundamental ability in human cognition~\cite{pelletier2004semantic}.
By breaking down complex questions into smaller, more manageable tasks, agents can better grasp the underlying intent and devise an executable plan to find the answers.
As we will demonstrate in section~\ref{sec:ablation_architecture}, generating all subtasks in the single-pass manner at the beginning~\cite{wang-etal-2023-plan,sun2023pearl} often leads to redundant or irrelevant ones. 
\texttt{DARA} incorporates and advocates iterative task decomposition, which generates one task at a time building on the results of the previous task.
During the process, the agent will decide if more tasks are needed based on whether the current \textit{s-expression} can get the answer to the given question.
This adaptive approach allows \texttt{DARA} to dynamically refine its focus and tailor subsequent subtasks, contributing to a more effective reasoning process.

\subsubsection{Task Grounding}\label{sec:task_grounding}
\paragraph{Skim-then-Deep-Reading Relation Selection} To ground the decomposed tasks against KGs, previous efforts~\cite{liu2020kbert,michihiro2022dragon} pre-train models on KG-augmented data.
However, such methods have limitations in scalability and coverage.
They can struggle to fully capture the extensive and diverse topics, entities, and relations present in large-scale KGs (e.g., Freebase has 44 million topics and 2.4 billion facts.)

To alleviate this issue, \texttt{DARA} introduces a two-stage relation selection method based on the recently emerging powerful natural language understanding ability of LLMs.
\texttt{DARA} first invokes $get\_relations$ or ${get\_relevant\_relations}$ to obtain relations and selects $n$ most likely candidates among them (\textit{Step 1.1: Skim Relations} in Figure~\ref{fig:overview}). Subsequently, descriptions of these selected candidates are obtained using the $get\_descriptions$ function (\textit{Step 1.1: Deep Reading Descriptions} in Figure~\ref{fig:overview}). By reading these descriptions, the most suitable one is selected (\textit{Step 1.1: Schema Selection} in Figure~\ref{fig:overview}).
Note that this approach eliminates the need for LLM agents to extensively learn the environment in advance, making it more adaptable and transferable across different scenarios. 

\paragraph{Logical Form Construction}\label{sec:lfc}
As depicted in Figure~\ref{fig:overview}, \texttt{DARA} automatically constructs the logical form of the current step based on the selected schema item and the preceding logical form (\textit{Step-level S-expression} in Figure~\ref{fig:overview}). 
Unlike ICL-based agents which have to acquire it via explanation in the prompt, the syntax of constructing logical forms is learned during fine-tuning. For the syntax and complex usage of logical forms (\textit{s-expression}), please refer to Appendix~\ref{sec:s-exp-syntax}.

\subsection{The Action Space} \label{sec:action space}
The action space $\mathcal{A}$ of \texttt{DARA} consists of a set of functions to interact with the KG. For Freebase, the space of $\mathcal{A}$ is as follows:
\begin{itemize} 
 \setlength\itemsep{-1mm}
    \item $get\_relations(entities, topk)$: This function returns top-$k$ 1-hop both incoming and outgoing relations for given entities. 
    \item $get\_relevant\_relations(task, topk)$: When no entities are mentioned in the question, this function retrieves top-$k$ relevant relations according to the decomposed task.
    \item $get\_classes(entities, topk)$: This function returns top-$k$ classes of given entities.
    \item $get\_relevant\_classes(task, topk)$: This function retrieves the top-$k$ relevant classes determined by the decomposed task.
    \item $get\_descriptions(schema)$: This function retrieves descriptions of input schema.
\end{itemize}
These actions are taken during task grounding to get necessary schema items from KGs.
The input argument \mbox{\textit{entities}} in above functions is either the entities appearing in the question or the intermediate results \textit{(s-expression)} during the reasoning.
The functions $get\_relevant\_relations$ and $get\_relevant\_classes$ utilize a bi-encoder retriever, specifically \textit{all-mpnet-base-v2}~\cite{reimers-gurevych-2019-sentence}, fine-tuned on \textit{<question, schema items>} in training data to fetch relevant relations or classes for questions lacking entities.
 In the case of functions $get\_relations$ and $get\_classes$, we utilize the same retriever to filter the top-$k$ candidates after getting schema items from the KGs. The rationale behind this choice is twofold: 1) it alleviates the selection burden on \texttt{DARA}; 2) it preserves context length for multiple rounds of reasoning.
 The input \textit{s-expression} of above functions are mapped to SPARQLs for querying the KG using APIs.

\section{Experiment Setup}
\subsection{Constructing Fine-tuning Data}
We create reasoning trajectories for \texttt{DARA} using GraphQ, WebQSP, and GrailQA. Pairs of \mbox{\textit{<question, s-expression}>} in those datasets are filtered based on criteria, including relation diversity, removal of duplicates, and inclusion of complex questions with at least two subtasks. Subsequently, we linearize the \textit{s-expression} and instruct GPT-4 to translate it into natural language.
In total, we obtain 768 instances with reasoning trajectories.
The construction details and the example of reasoning trajectories are provided in Appendix~\ref{sec: data creation}.

\subsection{Zero-shot Evaluation}
One of the preliminary challenges of interacting with KGs is the substantial volume of unseen relations, classes, and entities during training.
To simulate the real-world unseen scenario and for a fair comparison, we conduct zero-shot evaluation as introduced by GrailQA~\cite{yu2021grailqa}, referring to schemas in the test data not appearing in training data.
For GrailQA, the evaluation is performed on the development data since the golden entity linking results in the hidden test data are unavailable.
The resulting statistics of the filtered test data are presented in Table~\ref{tab:data size}.

\begin{table}[]
\centering
\resizebox{\linewidth}{!}{%
\setlength{\tabcolsep}{8pt}
\renewcommand{\arraystretch}{0.85}
\begin{tabular}{lcll}
\hline
          & \textbf{GrailQA} & \textbf{GraphQ}           & \textbf{WebQSP}        \\ \hline
Data Size & 3,274            & \multicolumn{1}{c}{1,229} & \multicolumn{1}{c}{56} \\ \hline
\end{tabular}%
}
\caption{Test data with unseen schemas of each dataset}
\label{tab:data size}
\end{table}

\subsection{Evaluation Metrics} 
We use two evaluation metrics: exact match which evaluates if the predicted and gold logical forms are semantically equivalent~\cite{yu2021grailqa} and F1 score based on predicted and gold answers.

\subsection{Baselines}
We compare \texttt{DARA} with three different baselines: ICL-based Agents, Fine-tuned Agents, and Bottom-up semantic parsers.
As \texttt{DARA} is a semantic-parsing LLM agent, we do not include retrieval-augmented generation methods in our baselines such as Think-on-Graph~\cite{sun2023think}, StructGPT~\cite{jiang2023structgpt}, which only measures the top-1 accuracy of answers (Hits@1).
The difference is detailed in Appendix~\ref{app:diff_rag}.

\paragraph{ICL-based Agents.} We use ICL-based LLM agents in AgentBench as baselines, which provides seven functions to interact with KG and perform logical reasoning.
GPT-4 and Llama-2-chat (70B) serve as the backbone LLMs.
\paragraph{Fine-tuned Agents.} To evaluate the efficacy of \texttt{DARA} against alternative fine-tuned LLM agents, we include models in AgentTuning~\cite{zeng2023agenttuning}, i.e., AgentLM-7B, AgentLM-13B as baselines. 
AgentLM 7B/13B fine-tuned Llama-2-7B and 13B on 324 reasoning trajectories generated by GPT-4 in the AgentBench framework.
As AgentLMs trained on different examples with \texttt{DARA}, to conduct a direct comparison, we also convert training data of \texttt{DARA} into AgentBench Framework and fine-tune the Llama-2-7B, called AgentBench-7B.
The only difference between AgentBench-7B and \texttt{DARA} (Llama-2-7B) lies in the reasoning framework being utilized, as all other factors remain consistent.
One example of the reasoning trajectory in AgentBench Framework is shown in Appendix Table~\ref{tab: exemplary reasoning trajectory of agentbench}.

\paragraph{Bottom-up semantic parser.}
Although comparing with bottom-up semantic parsers is not an apples-to-apples comparison, we still include two state-of-art models ArcaneQA~\cite{gu-su-2022-arcaneqa} and Pangu~\cite{gu-etal-2023-dont}.
This comparison allows for a comprehensive assessment of \texttt{DARA}'s performance in the realm of semantic parsing, shedding light on the effectiveness of language agents and potential advancements in the domain.

ArcaneQA employs a constraint decoding method to generate an executable sub-program at each step.
Pangu follows the enumeration-then-rank paradigm. At each step, it enumerates all possible subprograms and applies a discriminator to select top-k candidates with the highest scores.

\subsection{Implementation Details}
We full fine-tuned \texttt{DARA} based on Llama 2 7B/13B~\cite{hugo2023llama2}, CodeLlama 7B~\cite{rozi2023codellama}, Mistral-7B~\cite{jiang2023mistral} with 2-4 80GiB H100 and use one 40GiB A100 GPU for inference. The implementation details can be found in Appendix~\ref{app:implement}.

\begin{table*}[th]
\renewcommand{\arraystretch}{0.85}
\centering
\small
\setlength{\tabcolsep}{10pt}
\resizebox{0.88\linewidth}{!}{%
\begin{tabular}{@{}lllllll@{}}
\toprule
\multirow{2}{*}{\textbf{Model}} & \multicolumn{2}{c}{\textbf{GraphQ}} & \multicolumn{2}{c}{\textbf{GrailQA}} & \multicolumn{2}{c}{\textbf{WebQSP}}  \\ \cmidrule(l){2-7} 
 & \multicolumn{1}{c}{\textbf{EM}} & \multicolumn{1}{c}{\textbf{F1}} & \multicolumn{1}{c}{\textbf{EM}} & \multicolumn{1}{c}{\textbf{F1}} & \multicolumn{1}{c}{\textbf{EM}} & \multicolumn{1}{c}{\textbf{F1}} \\ \midrule
\multicolumn{7}{@{}l}{\textit{Off-the-shelf LLM Agent (in context learning)}} \\ \midrule
AgentBench (Llama-2-chat-70B) & 25.63 & 30.33 & 33.20 & 35.72 & 8.93 & 10.18  \\
AgentBench (GPT-4) & 53.86 & \multicolumn{1}{c}{64.48} & 63.56 & 65.89 & 25.00 & 32.09  \\ \midrule
\multicolumn{7}{@{}l}{\textit{Fine-tuned LLM Agent}} \\ \midrule
AgentLM-7B & 36.21 & 43.92 & 14.45 & 15.27 & 5.35 & 6.82  \\
AgentLM-13B & 44.34 & 54.35 & 52.72 & 55.01 & 12.50 & 14.98 \\
AgentBench-7B & 42.72 & 53.37 & 56.96 & 59.28 & 19.64 & 23.96 \\
\texttt{DARA} (this work) \\
\hspace{3mm} w/Llama-2-7B & 51.51 & \multicolumn{1}{c}{62.74} & 75.05 & 77.71 & \textbf{30.36} & \textbf{42.67} \\
\hspace{3mm} w/CodeLlama-7B & 51.83 & 63.15 & 75.26 & 78.61 & 21.43 & 32.15 \\
\hspace{3mm} w/Mistral-7B & \textbf{56.96} & \textbf{69.47} & 76.88 & 80.16 & 28.57 & 36.00 \\
\hspace{3mm} w/Llama-2-13B & 55.57 & \multicolumn{1}{c}{67.34} & 77.03 & 80.35 & \textbf{30.36} & 41.63\\
\midrule
\multicolumn{7}{@{}l}{\textit{Bottom-up Parser (full train data)}} \\ \midrule
ArcaneQA & 37.00 & 47.50 & 78.52 & 81.81 & 23.21 & 37.80 \\
Pangu (T5-base) & 56.06 & 66.70 & \textbf{88.30} & \textbf{91.76} & 19.64 & 32.64 \\
Pangu (T5-Large) & \textbf{55.57} & 67.21 & \multicolumn{1}{c}{$-^*$} & \multicolumn{1}{c}{$-^*$} & 23.21 & 36.72 \\ \bottomrule
\end{tabular}%
}

\caption{Overall zero-shot evaluation results in three different datasets. * indicates the trained T5-large on GrailQA is not provided in the GitHub repository of Pangu. All models use the golden entity linker. The full train data size of each dataset can be found in Appendix~\ref{tab:dataset stat}.}
\label{tab:main results}
\end{table*}



\section{Experiment Results}
\subsection{Overall Performance}
As presented in Table \ref{tab:main results}, \texttt{DARA} exhibit consistent superiority over both ICL-based agents and fine-tuned counterparts across three datasets.
It is crucial to highlight that alternative fine-tuned agents, such as AgentLM-7B, AgentLM-13B, and AgentBench-7B, exhibit inferior performance compared to GPT-4 agents.
For instance, on GrailQA, AgentLM-13B lags behind GPT-4 by 10.88\% while \texttt{DARA} (Llama-2-7B) and \texttt{DARA} (Llama-2-13B) outperform GPT-4 by 11.82\% and 14.46\% (F1), respectively.
Although AgentBench-7B is trained using the same data as \texttt{DARA}, it trails \texttt{DARA} (Llama-2-7B) by 18.57\%. 
These results serve as strong evidence of \texttt{DARA}'s effectiveness and superiority in handling KGQA tasks.
\texttt{DARA} shows flexibility with different back-end LLMs and exhibits improved performance commensurate with the capabilities of the selected model--opting for stronger LLMs results in enhanced performance.
Llama-2-13B can achieve better results than Llama-2-7B.
It is noteworthy that \texttt{DARA} (Mistral-7B) either outperforms or achieves comparable performance with \texttt{DARA} (Llama-2-13B) on GraphQ and GrailQA datasets.
\texttt{DARA} can also be utilized in the ICL fashion using prompting. When equipping GPT-4 with \texttt{DARA}, as demonstrated in Table~\ref{tab:icl-dara}, we show that \texttt{DARA} (GPT-4) achieved the best performance on 400 hardest questions with longest reasoning paths from GrailQA and GraphQA.

\begin{table}[htb]
\centering
\Huge
\renewcommand{\arraystretch}{1.3}
\resizebox{\columnwidth}{!}{%
\begin{tabular}{@{}lcccccc@{}}
\toprule
\multirow{2}{*}{\textbf{Model}}             & \multicolumn{2}{l}{\textbf{GraphQ-hard}} & \multicolumn{2}{l}{\textbf{GrailQA-hard}} & \multicolumn{2}{l}{\textbf{WebQSP}} \\ \cmidrule(l){2-7}
                   & EM   & F1       
                   & EM   & F1    
                   & EM   & F1 \\ \hline
\multicolumn{7}{@{}l}{\textit{ICL-based Agents}}                                                                     \\ \midrule
AgentBench (GPT-4) & 41.5         & 41.6        & 36.0         & 44.9      
&     25.0        &    32.1         \\
\texttt{DARA} (GPT-4)    &  \textbf{48.0}   &   \textbf{52.0}
    &  \textbf{51.5}           &   \textbf{61.6}           &     26.8        &   33.6                    \\ \midrule
\multicolumn{7}{@{}l}{\textit{Fine-tuned Agents}}                                                                    \\ \midrule
\texttt{DARA} (Llama-2-7B)  & 46.0         & 48.8        & 41.0         & 51.0        &    \textbf{30.4}         &    \textbf{42.67}       \\ \bottomrule
\end{tabular}%
}
\caption{The performance of \texttt{DARA} in the in-context learning fashion on difficult questions.}
\label{tab:icl-dara}
\end{table}

In comparison to bottom-up parsers, \texttt{DARA} showcases its prowess.
\texttt{DARA} (Mistral-7B and Llama-2-13B) outperforms both AcraneQA and Pangu (T5-large) on GraphQ and WebQSP.
Notably, \texttt{DARA} (Mistral-7B) outperforms AcraneQA by 21.97\% on GraphQ.
For GrailQA, while \texttt{DARA} lags behind Pangu (T5-large) by 11.41\%  (F1-score), it is important to note that Pangu utilized over 57 times more data (44,337 examples) for training, emphasizing the substantial efficiency of \texttt{DARA} in learning from a limited dataset (768 examples).
On WebQSP, \texttt{DARA} (Llama-2-7B) achieves the best performance, outperforming Pangu (T5-large) by 5.95\%.
However, an interesting observation is that the performance of all models is much lower than that of models trained on the other two datasets.
Pangu (T5-large) can achieve a 78.9\% F1 score on the whole test data of WebQSP~\cite{gu-etal-2023-dont} while it only achieves a 36.72\% F1 score on zero-shot evaluation.
\citet{shu2023data} has a similar observation on cross-dataset evaluation. The main reason is the different data construction methods and limited expressivity of current \textit{s-expression}.
We provide detailed analysis in Appendix~\ref{app:performance discreapnacy}.

\subsection{Deatailed Analysis on \texttt{DARA} Components}\label{sec:ablation_architecture}

\begin{table}[htb]
\Huge
\centering
\resizebox{1\columnwidth}{!} {
\renewcommand{\arraystretch}{1.3}
\begin{tabular}{l|c|c|c}
\toprule
                            & \textbf{GraphQ} & \textbf{GrailQA} & \textbf{WebQSP} \\ \midrule
\texttt{DARA}           & 62.74  & 77.71   & 42.67 \\ \midrule
w/ PD                    & 56.50(\textbf{6.24}$\downarrow$) & 71.22(\textbf{6.49}$\downarrow$)  & 28.28(\textbf{14.39}$\downarrow$)  \\
w/o SDR & 56.79(5.95$\downarrow$)  & 75.12(2.59$\downarrow$)   & 40.77(1.90$\downarrow$)  \\
w/o FT Rtr. & 61.29(1.45$\downarrow$) & 75.76(1.95$\downarrow$) & 36.18(6.49$\downarrow$) \\ \bottomrule

\end{tabular}%
}
\caption{Ablation study of different components of \texttt{DARA} (Llama-2-7B)  under F1 scores.}
\label{tab:ablation study}
\end{table}


\paragraph{The effectiveness of Iterative Task Decomposition (ITD).} As shown in Table~\ref{tab:ablation study}, when equipped with the single-pass pre-decomposition (PD), the performance drops more than 6\% on GraphQ, GrailQA and 14.39\% on WebQSP. 
After inspecting the errors made by PD, we found models cannot generate perfect decomposed tasks in a single-pass.
The generated tasks are often redundant or incorrect.
For the question: \textit{Which conference sponsor also sponsored the conference series with GridRepublic?} In the pre-decomposition approach, the decomposed tasks are as follows: \textit{Task 1 - Find the conference series with GridRepublic, Task 2 - Find the sponsor of the conference series with GridRepublic, and Task 3 - Find the conferences sponsored by the same sponsor}, in which \textit{Task 3} is unnecessary for answering the question.
The ITD method, on the other hand, would stop the iteration at Task 2 and return the answer.
These results underscore the importance of ITD for \texttt{DARA}, as it allows for dynamic task decomposition, leading to improved performance across various datasets.
\paragraph{The role of the Skim-then-Deep-Reading relation selection method.} The results presented in Table~\ref{tab:ablation study} demonstrate the significance of the skim-then-deep-reading relation (SDR) selection strategy. When this strategy is not employed, the performance experiences a decrease, especially in GraphQ (5.95\% drop).
To illustrate the impact, for the question: \textit{What vocal range is Pavarotti?} In the absence of the skim-then-deep-reading selection strategy, \texttt{DARA} selects the incorrect relation, \textit{music.artist.track}. However, with the selection strategy in place, the correct relation, \textit{music.opera\_singer.voice\_type} is selected, whose description contains information about the vocal range.
These results highlight the critical role that the skim-then-deep-reading relation selection strategy plays in enhancing \texttt{DARA}'s performance especially when the literal meaning of relations cannot express the needed information.

\paragraph{The fine-tuned retriever for schema filtering.} To better select the relations or classes, we leverage a fine-tuned retriever to filter the top five candidates for the \texttt{DARA} (Section~\ref{sec:action space}).
As shown in Table~\ref{tab:ablation study}, the fined-tuned retriever (\mbox{FT Rtr.}) contributes to the improvement in model performance. Here, an off-the-shelf retriever \textit{all-mpnet-base-v2}\footnote{\url{https://www.sbert.net/docs/pretrained_models.html}} is used.
On WebQSP, without the trained retriever, the performance drops by 6.49\%.
\subsection{Error Analysis}\label{sec:error analysis}
To have a concrete understanding of the utility of \texttt{DARA}, we compare its reasoning trajectories with those of ICL-based agents (GPT-4, Llama-2-70B chat) and fine-tuned AgentBench-7B. The common errors among these agents were identified during the grounding phase.
AgentBench-7B exhibits a deficiency in schema understanding, selecting wrong relations.
In contrast, GPT-4 showcases a superior grasp of schema understanding.
However, both agents tend to prematurely terminate tasks when reaching a CVT node with n-ary relations\footnote{CVT is the mediator in Freebase so the agent needs a further step to get the final answer.}.
\texttt{DARA}, on the other hand, effectively identifies the CVT node through its skim-then-deep-reading relation selection component.
Moreover, the unthoughtful designed action space in the AgentBench framework not only limits the agents' capabilities for addressing questions that involve class information but also hinders their performance in handling queries lacking explicit entities.
Furthermore, for ICL-based LLM agents, the challenge of instruction-following persists.
For example, Llama-2-chat (70B) encounters challenges in following the reasoning pipeline provided in the demonstration, leading to failure function calls.
For fine-tuned AgentBench-7B, without an explicit task decomposition module like \texttt{DARA}, it struggles to obtain effective high-level guidance for some complex questions.
Comparison examples between those agents are provided in Appendix~\ref{app:error anlaysis}.

\subsection{The Quality of GPT-4 Generated Data}\label{sec: reasoning trajectories}
\begin{figure}
    \centering
    \begin{subfigure}{0.45\columnwidth}
        \includegraphics[width=\textwidth]{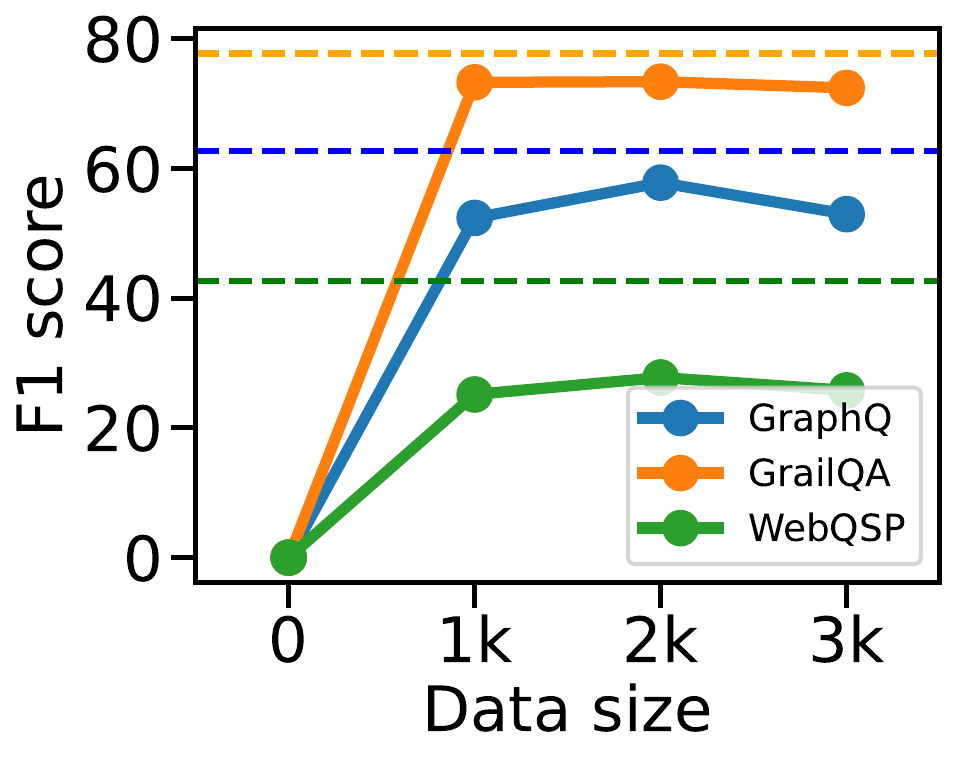}
        \caption{Llama-2-7B}
    \end{subfigure}
    \begin{subfigure}{0.45\columnwidth}
        \includegraphics[width=\textwidth]{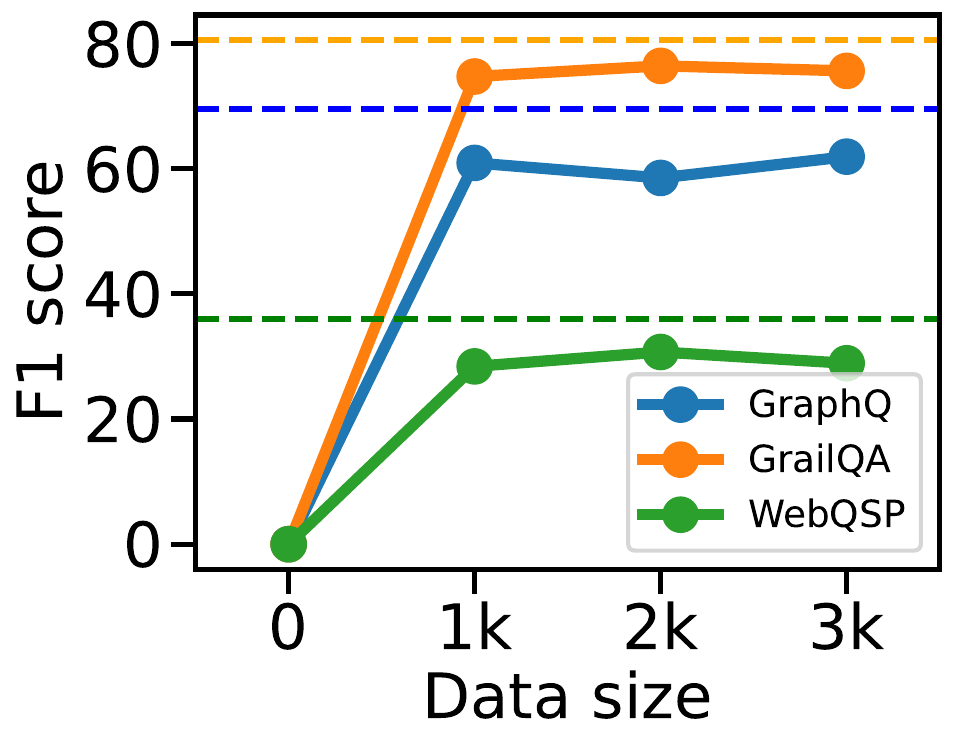}
        \caption{Mistral-7B}
    \end{subfigure}
    \caption{The performance of \texttt{DARA} on three datasets trained with reasoning trajectories generated by GPT-4. - - - (dashed line) represents the performance of 768 human-rectified examples.}
    \label{fig:gpt4 generated data}
\end{figure}

The GPT series has demonstrated exceptional performance in data generation and annotation~\cite{gilardi2023chatgpt,wang-etal-2023-self-instruct,xu2023wizardlm}.
We investigate whether reasoning trajectories converted by GPT-4 suffice for language agents fine-tuning in KGQA.
We fine-tuned \texttt{DARA} on GPT-4 converted trajectories on both Llama-2-7B and Mistral-7B.
As illustrated in Figure~\ref{fig:gpt4 generated data}, \texttt{DARA} with different back-end LLMs exhibits a similar pattern. Despite data size increasing from 1k to 3k, the performance on three datasets always lags behind that of 768 human-verified examples.
Upon closer inspection of the generated data, we identified the following issues.
GPT-4 struggles to follow the structured output format, i.e., confusing `step' and `task' and misinterprets the intent of the given logical form, thus generating wrong tasks. 
Furthermore, the schema design of Freebase (e.g. CVT) increases the difficulty of comprehension.
AgentTuning~\cite{zeng2023agenttuning} corroborates our findings, reporting a low success rate when utilizing GPT-4 to generate reasoning trajectories across six diverse environments, ranging from 0.52\% on Mind2Web to 35.2\% on AlfWorld.
These collective observations highlight a potential future direction for automatically generating high-quality intermediate trajectories for language agents in various environments, particularly when GPT-4's performance is constrained.

\subsection{The Running Expenses of Models}\label{sec:expense}
In Table~\ref{sec:expense}, we analyze the running expenses of various models on 4,559 test examples. Fine-tuned agents, such as \texttt{DARA}, outperform ICL-based LLM agents like GPT-4 and Llama-2-Chat (70B) in both cost and speed. GPT-4 stands out as the most expensive, with costs sixty times higher and speed four times slower than \texttt{DARA} (Llama-2-7B). Pricing information for the GPT-4 experiment is available on the Azure webpage\footnote{\url{https://azure.microsoft.com/en-us/pricing/details/cognitive-services/openai-service/}}. To assess the price of GPUs used for Llama-2, we refer to Replicate.com\footnote{\url{https://replicate.com/pricing}}. While the cost may be affected by different factors (e.g. GPU service providers, optimized inference methods), the affordability and efficiency of fine-tuned agents remain evident.

\begin{table}[]
\centering
\resizebox{\columnwidth}{!}{%
\renewcommand{\arraystretch}{0.8}
\begin{tabular}{ll|c}
\toprule
\multirow{2}{*}{Model} & \multicolumn{2}{c}{Cost} \\ \cmidrule{2-3} & Money  & Time (hours)  \\ \midrule
\small{AgentBench (Llama-2-chat (70B))} & \small{\$1,208}    & \small{\textbf{120.07}}    \\ \midrule
\small{AgentBench (GPT-4)} & \small{\textbf{\$1,276}} & \small{32.07} \\ \midrule
\small{\texttt{DARA} (Llama-2-7B)}            & \small{\$20.51}        & \small{5.01}          \\ \midrule
\small{\texttt{DARA} (Llama-2-13B)}           & \small{\$31.90}          & \small{7.72}           \\ \bottomrule
\end{tabular}%
}
\caption{The total cost of different models on 4,559 examples during inference.}
\label{tab:cost}
\end{table}

\section{Conclusion}
In this paper, we present \texttt{DARA}, a fine-tuned LLM agent for KGQA.
Experimental results highlight the superior performance of \texttt{DARA} with various backend LLMs, compared to ICL-based LLM agents as well as alternative fine-tuned language agents.
Moreover, \texttt{DARA} proves to be more cost-effective and time-efficient than ICL-based agents.
Additionally, our findings reveal challenges faced by GPT-4 in converting structured logical forms into natural language for reasoning trajectory construction.
In the future, we will center on developing methods to generate high-quality reasoning trajectories for LLM agent fine-tuning in KGs and other symbolic environments where GPT-4 faces difficulties.

\section{Acknowledgements}
We thank Kexin Wang, Hiba Arnaout, Imbesat Hassan Rizvi, Haritz Puerto, and Tilek Chubakov for their valuable discussion and feedback. This work was supported by the Konrad Zuse School of Excellence in Learning and Intelligent Systems (ELIZA) through the DAAD programme Konrad Zuse Schools of Excellence in Artificial Intelligence, sponsored by the Federal Ministry of Education and Research.

\section*{Limitations}
Although \texttt{DARA} shows superiority over ICL-based and other fine-tuned LLM agents, there are several limitations that call for further improvement.
First, DARA lacks the error-correcting ability, which means it cannot correct itself when erroneously decomposing the question or selecting incorrect relations.
Humans excel in fixing errors based on the current observation and the final goal.
Exploring approaches like Reflextion~\cite{Shinn2023ReflexionLA} could enhance error-correcting capabilities, although it is important to note that such methods currently work effectively only with very powerful LLMs, such as GPT-4.
Second, we performed evaluations in the zero-shot setting due to the high cost of commercial LLMs.
To create fine-tuning data, we use GPT-4 to translate 768 golden logical forms to natural language, which increases the complexity of the method.
Third, following previous research, we only evaluate it on popular datasets using Freebase as the backend.
For other knowledge graphs such as Wikidata, the current intermediate representation \textit{s-expression} can not be directly transferred. 
Generalizing \texttt{DARA} to other knowledge graphs will be our future work.

\bibliography{anthology,custom}

\appendix \label{sec:appendix}
\section{Reasoning Trajectories Creation}\label{sec: data creation}
For <question, \textit{s-expression}> pairs in the original dataset, we linearized the \textit{s-expression} and break down it into several tasks according to logical operations (refer to Appendix~\ref{sec:logical operations}). 
These tasks serve as golden decomposed tasks.
To convert them into natural language, we prompt GPT-4 coupled with the corresponding question using the prompt in Appendix~\ref{sec:decompose_prompt}.
For the reasoning trajectories of the skim-then-deep-reading relation selection component, we automatically create them using the schema items in golden \textit{s-expression} and functions in Section~\ref{sec:action space}.
Then the data is inspected and rectified by humans.

\subsection{Dataset Quality Inspection}\label{sec:data quality}
Fine-tuning LLMs with large but very noisy data contributes little to performance improvement~\cite{zhou2023lima, lee2023platypus}.
We manually inspect selected data from the original dataset.
Several issues are identified: unnatural expressions, redundant logical forms, ambiguous questions, and inconsistencies between questions and logical forms.
GrailQA is a large-scale dataset aiming at evaluating different level generalization abilities of KGQA models.
Among 500 sampled data points, roughly 15\% contained errors.
The main issue is the wrong directionality of relations. 
In the case where the question is asking about \textit{operating systems that include 386bsd}, the logical form is searching for operating systems belonging to 386sd \textit{(JOIN computer.operating\_system.parent\_os 386BSD)}.
Another common issue is improper comparative operations. For instance, the question is asking about the comparison \textit{more than} while \textit{ge (greater than or equal to)} is adapted in logical forms. Besides, some errors involve redundant components within the query, which is unnecessary to answer the question.
GraphQ is a medium-sized dataset where questions are generated by expert annotators according to the query graph.
It has multiple paraphrased questions in the same logical form. We only retain one question for each logical form.
WebQSP is another medium-sized dataset comprising questions sourced from Google query logs. This dataset includes implicit entities not mentioned in the questions and complex logical operations. Furthermore, it contains many ambiguous/open questions. 
For example, \textit{what did Stephen Hawking become famous for?} is an open question. The answer could be his profession, book, discovery, etc.
The data statistics of these three datasets are shown in Table~\ref{tab:dataset stat}.
\begin{table}[htb]
\centering

\begin{tabular}{lccc}
\hline
                  & \textbf{GraphQ} & \textbf{GrailQA} & \textbf{WebQSP} \\ \hline
\textbf{Training} & 2,381           & 44,337           & 3,098           \\ \hline
\textbf{Dev}      & -               & 6,763            & -               \\
\textbf{Test}     & 2,395           & 13,231           & 1,639           \\ \hline
\end{tabular}%

\caption{Original Dataset Statistics}
\label{tab:dataset stat}
\end{table}

\begin{table*}
\centering
\Huge
\renewcommand{\arraystretch}{1.1}
\resizebox{\textwidth}{!}{%
\begin{tabular}{lll}
\hline
\textbf{Logical Operation}   & \textbf{Returns}             & \multicolumn{1}{c}{\textbf{Descriptions}}                                            \\ \hline
(AND u1 u2)         & a set of entities               & AND function returns the intersection of two arguments                      \\ \hline
(COUNT u)           & a singleton set of integer      & COUNT function returns the cardinality of the argument                      \\ \hline
(R b)               & a set of (entity entity) tuples & R function reverse each binary tuple (x, y) in the input to (y, x)          \\ \hline
(JOIN b u)          & a set of entities               & Inner join based on items in u and the second element of items in b         \\ \hline
(JOIN b1 b2) &
  a set of (entity, entity) tuples &
  \begin{tabular}[c]{@{}l@{}}Inner join based on the first element of items in b2 and the second\\ element of items in b1\end{tabular} \\ \hline
(ARGMAX/ARGMIN u b) & a set of entities               & Return x in u such that (x, y) is in b and y is the largest / smallest      \\ \hline
(LT/LE/GT/GE b n)   & a set of entities               & Return all x such that (x, v) in b and v \textless /  / \textgreater /  n \\ \hline
\end{tabular}%
}
\caption{The definition and syntax of \textit{s-expression}}
\label{tab:s_expression_exp}
\end{table*}
\subsection{Logical Operations}\label{sec:logical operations}
The common logical operations are as follows:
\begin{itemize}
    \item Superlative (ARGMIN/ARGMAX): it is used to get entities with the maximum value of a given attribute.
    \item Intersection (AND): it is used to get an intersection between two sets of entities.
    \item Comparative (LT/LE/GE/GE): it is used to filter properties according to numerical constraints. 
    \item Count (COUNT): it is used to count the number of a set of entities.
    \item Projection (JOIN): This operation is used to get the other side entity of a triplet via the relation.
\end{itemize}
We break down the logical form into subtasks according to the above logical operations.
For projection operation, it may be used due to the KG structure rather than the question.
For such cases, we do not break it down.

\subsection{Logical form used in \texttt{DARA}}\label{sec:s-exp-syntax}
We use s-expression~\cite{yu2021grailqa} as the logical form to represent SPARQL. The definition and syntax of logical operations are as shown in Table~\ref{tab:s_expression_exp}.

\subsection{Statistics of training reasoning trajectories}
As shown in Table~\ref{tab:dara data}, we utilize 768 reasoning trajectories to train \texttt{DARA}.
\begin{table}
\centering
\resizebox{0.9\columnwidth}{!}{%
\setlength{\tabcolsep}{4pt}
\begin{tabular}{lclll}
\hline
  & \textbf{GrailQA} & \textbf{GraphQ} & \textbf{WebQSP}    & \textbf{Total number}     \\ \hline
  & 426       & \multicolumn{1}{c}{193} & \multicolumn{1}{c}{149} & \multicolumn{1}{c}{768} \\
          
\hline
\end{tabular}%
}
\caption{The size of curated training data from different datasets with reasoning trajectories.}
\label{tab:dara data}
\end{table}

\clearpage

\section{Difference between RAG methods and Semantic-parsing methods}\label{app:diff_rag}
Retrieval-augmented generation method (RAG) such as Think-on-Graph, StructGPT as discussed in the related work (Section~\ref{sec:related_work}), which uses Hits@1 as the metric to measure the top-1 accuracy, which does not consider the recall. Semantic-parsing methods (e.g. \texttt{DARA}, Pangu, ArcaneQA, AgentTuning) derive logical forms to query the KGs, which use the F1-score to measure recall and precision of answers obtained from logical forms. Given these fundamental differences in goals and metrics, drawing direct comparisons between these methodologies may not fully capture their respective strengths and limitations.
RAG methods rely on the LLM to synthesize responses based on extracted triplets and its internal knowledge, rather than directly querying a KG. For a question with multiple answers, ToG cannot guarantee getting all answers while semantic parsing methods (e.g. DARA) can get all answers from the KGs by executing the derived logical forms.
To give a quantitative analysis, we compare ToG (GPT-4) with 
\texttt{DARA} (llama 2-7B, mistral-7B) on 400 GrailQA test questions with multiple answers using both hit@1 and F1-score. 
As shown in Table~\ref{tab:diff_rag_parser}, \texttt{DARA} outperforms ToG in the tested 400 examples. We can see ToG has a quite low F1 score as we discussed before. The Hit@1 score is also lower than DARA. Evaluating ToG on the F1 score might seem unfair, given its design for high accuracy in single-answer scenarios. Conversely, using hit@1 does not fully capture the recall and precision of logical forms generated by \texttt{DARA}.
\begin{table}[htb!]
\centering
\resizebox{0.9\columnwidth}{!}{%
\begin{threeparttable}
\begin{tabular}{@{}lll@{}}
\toprule
Models & Hits@1 (\%) & F1-score (\%) \\ \midrule
ToG (GPT-4) & 65.2\tnote{*} & 7.7 \\
DARA (Llama 2 7B) & 70.5 & 69.6 \\
DARA (Mistral) & 75.8 & 72.6 \\ \bottomrule
\end{tabular}%
\begin{tablenotes}
    \item[*] Follow the original paper of ToG, this result excludes questions without entities since ToG cannot solve them. For DARA, all questions are used to calculate the metrics.
\end{tablenotes}
\end{threeparttable}
}

\caption{Comparsion between ToG with \texttt{DARA} on 400 questions in GrailQA with multiple answers}
\label{tab:diff_rag_parser}
\end{table}

\section{Implementation Details}\label{app:implement}
\paragraph{Training} We fine-tuned Llama-2 and CodeLlama using 2-4 80GiB H100 for 10 epochs using
2e-5 learning rate with a 3\% warm-up ratio and 4 batch size. Deepspeed is utilized to accelerate the training process.
For Mistral-7B, the learning rate is 2e-6 learning rate with a 10\% warm-up ratio.

\paragraph{Inference}
To do inference on Llama 7B/13B, we use one 40GiB A100 GPU while 2 80GiB A180 GPUs are used to test Llama-2-chat-70B.
We call Azure OpenAI service API for GPT-4 inference.

\section{Prompt and reasoning trajectories}
\subsection{Prompt used to convert logical forms to natural languages}\label{sec:decompose_prompt}
We present the prompt and demonstrations used to convert broken-down logical forms into natural languages in Table ~\ref{tab:prompt for conversion}.

\begin{table*}
\centering
\begin{tabularx}{\textwidth}{X} 
\toprule
\textbf{User}: You are knowledgeable about Freebase. You need to describe what is the intent of logical forms in the given subtasks and steps. To find the answer to the question 'Which type of dish common in manado cuisine contains sugars? The linked entities are sugars (m.06x4c), and manado cuisine (m.0102k5v9).', it has the following subtasks.
Note: Some tasks may have multiple steps due to the schemas of the KG. \\
Task 1: Step1:(JOIN food.dish.ingredients m.06x4c)\\
Task 2: Step1:(JOIN (R dining.cuisine.dishes) m.0102k5v9)\\
Task 3: Step1:(AND task1 task2)\\
Task 4: Step1:(JOIN food.type\_of\_dish.dishes task3)\\
To help you understand the relations used in the above steps, we provide the following descriptions of them. 1. the incoming relation 'food.dish.ingredients', which describes please enter the main ingredients usually included in this dish. The type of its head entity is 'food.dish' (A dish is a food prepared and presented in a certain way. These are foods that are ready to be served and eaten as a meal itself, or part of a larger meal). 2. the outgoing relation 'dining.cuisine.dishes', which describes the dishes of cuisine. The type of its tail entity is 'food.dish' (A dish is a food prepared and presented in a certain way. These are foods that are ready to be served and eaten as a meal itself, or part of a larger meal). 3. the incoming relation 'food.type\_of\_dish.dishes', which describes the dishes of type of dish. The type of its head entity is 'food.type\_of\_dish' (This type is used to classify dishes into broad categories, such as appetizers, soups, desserts, etc.).
Please describe the intent of the above tasks and steps.\\
\textbf{Assistant}: To solve this question, we can break it down into the following tasks:\\ 
Task 1: Determine the dishes that contain sugars.\\
- Step 1.1: use the incoming relation 'food.dish.ingredients' of surgars with the operator JOIN to find dishes whose ingredients contain sugars.\\
Task 2: Identify the dishes in Manado cuisine.\\
- Step 2.1: we can use the outgoing relation 'dining.cuisine.dishes' of manado cuisine to find the dishes in Manado cuisine.\\
Task 3: Find the common dishes from Task 1 and Task 2.\\
- Step 3.1: we can use the operator AND to get the common dishes from Task 1 and Task 2.\\
Task 4: Determine the type of dish among the common dishes.\\
- Step 4.1: to get the type of dish, we use the the incoming relation 'food.type\_of\_dish.dishes' of those entities obtained from Task 3 to get the final answer.\\
\textbf{User}: To find the answer to the question 'which academic was an dashiell that was born in japan? The linked entities are Japan (m.03\_3d), dashiell (m.03m3mgq).', it has following tasks.\\
Each task may have multiple steps due to the schemas of the KG.\\
Task 1: Step1:(JOIN people.person.place\_of\_birth m.03\_3d)\\
... \\
Please describe the intent of the above tasks and steps. \\
\textbf{Assistant}: To solve this question, we can break it down into following tasks:\\
Task1: find people who were born in Japan.\\
...\\
\textbf{User}: To find the answer to the question '\{question\}', it has following tasks.\\\{subtasks\}\\To help you understand the relations used in above steps, we provide the following descriptions of them.\\\{description\}\\Please describe the intent of above tasks and steps. \\
\bottomrule
\end{tabularx}
\caption{The prompt used to convert linearized logical forms to natural language.}
\label{tab:prompt for conversion}
\end{table*}

\subsection{A exemplary reasoning trajectory of \texttt{DARA}}~\label{sec:example of rt}
As shown in Table~\ref{tab: exemplary reasoning trajectory}, we provide the complete reasoning trajectory of \texttt{DARA}, which includes iterative task decomposition and the two-stage skim-then-deep-reading relation selection method.

\begin{table*}
\begin{tabularx}{\textwidth}{X} 
\toprule
\textbf{User:} For a given question, your task is to parse the question into a correct logical form (s-expression) which could be executed over a KG to get the answer. To write the s-expression, you need to decompose the question into subtasks and solve them step-by-step. To get necessary schemas (i.e., relations or classes) for s-expression writing, you can use the following functions to interact with the KG.\\
get\_relations(expression): return all relations of those entities represented by the expression.\\
get\_classes(expression): return all class types of those entities represented by the expression.\\
get\_relevant\_relations(thought): retrieve relevant relations from the KG according to your thought.\\
get\_relevant\_classes(thought): retrieve relevant classes from the KG according to your thought.\\
get\_descriptions(candidate): get description of candidate schemas.
The question is what olympic games did ronny represent as a mascot? The linked entity is Ronny (m.04dwjbg). \\
\textbf{Agent:} The given question can be decomposed into the following subtasks: \# Task 1: Find the Olympic games that Ronny represented as a mascot. \\
\#\# Step 1.1: \\
\#\#\# Action 1.1.1: get\_relations(m.04dwjbg)\\
\#\#\# Obs 1.1.1: m.04dwjbg has following relations. The outgoing relations are [olympics.olympic\_mascot.olympic\_games, kg.object\_profile.prominent\_type, common.topic.notable\_for, common.topic.notable\_types, type.object.name]. The incoming relations are []. \\
\#\#\# Thought 1.1.1: From the above relations, olympics.olympic\_mascot.olympic\_games (outgoing), kg.object\_profile.prominent\_type (outgoing) are the two most likely relations to finish Task 1. To select the correct one, I need to check their underlying meaning.\\
\#\#\# Action 1.1.2: get\_descriptions(olympics.olympic\_mascot.olympic\_games (outgoing), kg.object\_profile.prominent\_type (outgoing)) \\
\#\#\# Obs 1.1.2: 1. the outgoing relation 'olympics.olympic\_mascot.olympic\_games', which describes the Olympic games for which this was a mascot. The type of its tail entities is 'olympics.olympic\_games' (This type is for any modern Olympic Games, summer or winter.). 2. the outgoing relation 'kg.object\_profile.prominent\_type', which describes the prominent type of object profile. The type of its tail entities is 'prominent\_type' (prominent type). \\
\#\#\# Thought 1.1.2: From the explanations, use the outgoing relation 'olympics.olympic\_mascot.olympic\_games' of Ronny with the operator JOIN to find the Olympic games that Ronny represented as a mascot.\\
\#\#\# S-exp-1.1: (JOIN (R olympics.olympic\_mascot.olympic\_games) m.04dwjbg)\\

\#\# S-exp-1: (JOIN (R olympics.olympic\_mascot.olympic\_games) m.04dwjbg)\\
\# Final s-exp:
(JOIN (R olympics.olympic\_mascot.olympic\_games) m.04dwjbg)\\

\bottomrule
\end{tabularx}
\caption{An exemplary reasoning trajectory of \texttt{DARA} for fine-tuned agents.}
\label{tab: exemplary reasoning trajectory}
\end{table*}

\begin{table*}
\begin{tabularx}{\textwidth}{X} 
\toprule
\textbf{User:} You are an agent that answers questions based on the knowledge stored in a knowledge base. To achieve this, you can use the following tools to query the KB.
1. get\_relations(variable: var) -> list of relations \\
A variable can be either an entity or a set of entities (i.e., the result of a previous query). This function helps to navigate all relations in the KB connected to the variable, so you can decide which relation is the most useful to find the answer to the question.
A simple use case can be 'get\_relations(Barack Obama)', which finds all relations/edges starting from the entity Barack Obama.
The argument of get\_relations should always be an entity or a variable (e.g., \#0) and not anything else.

2. get\_neighbors(variable: var, relation: str) -> variable \\
Given a variable, this function returns all entities connected to the variable via the given relation. Note that, get\_neighbors() can only be used after get\_relations() is used to find a set of viable relations.
A simple use case can be 'get\_neighbors(Barack Obama, people.person.profession)', which returns the profession of Obama in Freebase.

3. intersection(variable1: var, variable2: var) -> variable \\
Given two variables, this function returns the intersection of the two variables. The two variables MUST be of the same type!

4. get\_attributes(variable: var) -> list of attributes \\
This function helps to find all numerical attributes of the variable. Please only use it if the question seeks for a superlative accumulation (i.e., argmax or argmin).

5. argmax(variable: var, attribute: str) -> variable \\
Given a variable, this function returns the entity with the maximum value of the given attribute. It can only be used after get\_attributes() is used to find a set of viable attributes.
A simple use case can be 'argmax(variable, age)', which returns the oldest entity belonging to the variable.

6. argmin(variable: var, attribute: str) -> variable \\
Given a variable, this function returns the entity with the minimum value of the given attribute. It can only be used after get\_attributes() is used to find a set of viable attributes.
A simple use case can be 'argmin(variable, age)', which returns the youngest entity belonging to the variable.

7. count(variable: var) -> int \\
Given a variable, this function returns the number of entities belonging to the variable.

After a variable is produced along the process, you need to judge whether a variable is the final answer to the question. Each variable is represented as an id starting from 0. For example, \#0 is the first variable, \#1 is the second variable, and so on.
Once you find the answer, respond with 'Final Answer: \#id', where id is the id of the variable that you think is the final answer. For example, if you think \#3 is the final answer, you MUST respond with 'Final Answer: \#3'.

You can only take ONE action at a time!! After you get the observation from its execution, you can take another action. You can take at most 15 actions to find the answer to the question.
 A question is  which group does great comet of 1843 belong to?
Entities: [great comet of 1843] \\
\textbf{Agent:} Thought: Determine the group that the great comet of 1843 belongs to. \\
Action: get\_relations(great comet of 1843) \\
Observation: [astronomy.star\_system\_body.star\_system, astronomy.comet.comet\_group, astronomy.celestial\_object.category] \\

Thought: Use the outgoing relation 'astronomy.comet.comet\_group' of the great comet of 1843 (m.0595vt) with the function get\_neighbors to find the group that it belongs to. \\
Action: get\_neighbors(great comet of 1843, astronomy.comet.comet\_group) \\
Observation: variable \#0, which are instances of astronomy.comet\_group \\
Final Answer: \#0 \\
\bottomrule
\end{tabularx}
\caption{An exemplary reasoning trajectory of AgentBench Framework}
\label{tab: exemplary reasoning trajectory of agentbench}
\end{table*}

\onecolumn
\begin{longtable}[c]{p{\textwidth}}
\hline
\toprule
\textbf{User:} You are an agent that answers questions based on the knowledge stored in a knowledge base.\\ For a given question, you find the answer via iteratively task decomposition.
For each task, you need to complete it using the provided functions. After completing the task, you need to decide if further tasks are needed baed on the current results. The provided functions are: \\
1. get\_relations(variable: var) -> list of relations. A variable can be either an entity (represented by the mid) or a set of entities (i.e., the result of a previous query). This function helps to navigate all relations in the KB connected to the variable, so you can decide which relation is the most useful to find the answer to the question. A simple use case can be 'get\_relations(g.013c2d)', which finds all relations/edges starting from the entity g.013c2d. The argument of get\_relations should always be an entity or a variable (e.g., \#0) and not anything else.\\
2. get\_neighbors(variable: var, relation: str) -> variable Given a variable, this function returns all entities connected to the variable via the given relation. Note that, get\_neighbors() can only be used after get\_relations() is used to find a set of viable relations. A simple use case can be 'get\_neighbors(m.013c1m, people.person.profession)', which returns the profession of m.013c1m in Freebase.\\
3. intersection(variable1: var, variable2: var) -> variable. Given two variables, this function returns the intersection of the two variables. The two variables MUST be of the same type!\\
4. get\_relevant\_relations(task: str) -> list of relations. Given the decomposed task, this function helps to retrieve related relations. Note that, this function can only be used when no entity is mentioned in the task. Otherwise, get\_relations() should be used to get relations of the entities.\\
5. argmax(variable: var, attribute: str) -> variable. Given a variable, this function returns the entity with the maximum value of the given attribute. A simple use case can be 'argmax(variable, age)', which returns the oldest entity belonging to the variable. \\
6. argmin(variable: var, attribute: str) -> variable. Given a variable, this function returns the entity with the minimum value of the given attribute. A simple use case can be 'argmin(variable, age)', which returns the youngest entity belonging to the variable.\\
7. count(variable: var) -> int. Given a variable, this function returns the number of entities belonging to the variable.\\
8. get\_classes(variable:var) -> list of classes. A variable is a set of entities (i.e., the result of a previous query). This function returns the classes of given entities, which can be used to filter entities of certain classes.\\
9. get\_relevant\_classes(task: str) -> list of classes. Given the deomposed task, this function returns related relations of a decomposed task when there is no entitiy in the task. Otherwise, get\_classes() should be used.\\
10. lt(atrribute: str, value: float) -> variable. Given a attribute and a numerical value, this function returns the entity whose attribute is less than (lt) the given value. This function can only be used after get\_relations() or get\_relevant\_relations(). Similarly, le(attribute:str, value:float) is to find entities whose attribute is less than or equal to (le) the given value. ge(relation:str, value:float) is to find entities whose attribute is greater than or equal to (get). gt(relation:str, value:float) is to find entities whose attribute is greater than (gt) the given value.\\ 11.get\_descriptions(relation\_list:str) -> descriptions of given relations. This function returns descriptions of the given relation list. The input relation list should be seperated by comma. After getting relations using get\_relations or get\_relevant\_relations, if you are unsure which relation to select, select several most possible candidates and use this function to get descriptions of them to help you make the final decision.\\ 
12. get\_attributes(variable: var) -> list of attributes. This function helps to find all numerical attributes of the variable. Please only use it if the question seeks for a comparative accumulation (i.e., argmax/argmin, le/lt/ge/gt) or numerical attributes of entities.\\
13. get\_relevant\_attributes(task: str) -> list of attributes. This function finds relevant attributes of a task. \\ Note it can only be used when there is no entity in the question! Otherwise, get\_attributes should be used. After a variable is produced along the process, you need to judge whether a variable is the final answer to the question. Each variable is represented as an id starting from 0. For example, \#0 is the first variable, \#1 is the second variable, and so on. Once you find the answer, respond with 'Final Answer: \#id', where id is the id of the variable that you think is the final answer. For example, if you think \#3 is the final answer, you MUST respond with 'Final Answer: \#3' at the end to finish the question. You can only take ONE action at a time!! After you get the observation from its execution, you can take another action. You can take at most 15 actions to find the answer to the question. \\Here is a demo that you can use to get started. \\
Question: which bi-propellant rocket engines use unsymmetrical dimethylhydrazine and have a dry mass of less than 980.0? The linked entity is Unsymmetrical dimethylhydrazine (m.017q1y).\\
The given question can be decomposed into the following subtasks:\\
\# Task 1: Find bi-propellant rocket engines that use unsymmetrical dimethylhydrazine as fuel. \\
\#\# Step 1.1:\#\#\# Action 1.1.1: get\_relations(m.017q1y) \\
\#\#\# Observation 1.1.1: [spaceflight.rocket\_engine\_fuel.rocket\_engines, chemistry.chemical\_compound.average\_molar\_mass]. \\
\#\#\# Thought 1.1.1: From the above relations, I use the outgoing relation 'spaceflight.rocket\_engine\_fuel.rocket\_engines' of Unsymmetrical dimethylhydrazine with the function get\_neighbors to find bi-propellant rocket engines that use this fuel. \\
\#\#\# Action 1.1.2: get\_neighbors(m.017q1y, spaceflight.rocket\_engine\_fuel.rocket\_engines) \\
\#\#\# Observation 1.1.2: variable \#0, which are instances of spaceflight.bipropellant\_rocket\_engine.\\
\# Task 2: Find bi-propellant rocket engines with a dry mass less than 980.0.\\
\#\# Step 2.1:\#\#\# Action 2.1.1: get\_relevant\_relations(Find bi-propellant rocket engines with a dry mass less than 980.0.) \\
\#\#\# Observation 2.1.1: The relevant relations are spaceflight.rocket\_engine.dry\_mass, spaceflight.rocket\_engine\_fuel.rocket\_engines, spaceflight.bipropellant\_rocket\_engine.wet\_mass, spaceflight.rocket\_engine\_cycle.rocket\_engines. \\
\#\#\# Thought 2.1.1: we can use spaceflight.rocket\_engine.dry\_mass with the function lt to find bi-propellant rocket engines with a dry mass less than 980.0. \\
\#\#\# Action 2.1.2: lt(spaceflight.rocket\_engine.dry\_mass, 980.0^^http://www.w3.org/2001/XMLSchema\#float)\#\#\# Observation 2.1.2: \#1, which are instances of spaceflight.bipropellant\_rocket\_engine.\\
\# Task 3: Find the bi-propellant rocket engines that satisfy both Task 1 and Task 2.\\
\#\# Step 3.1:\#\#\# Thought 3.1.1: we can use the function intersection to get the common bi-propellant rocket engines from Task 1 and Task 2. The final answer would be the set of bi-propellant rocket engines that use unsymmetrical dimethylhydrazine and have a dry mass less than 980.0. This should be the final s-expression. \\
\#\#\# Action 3.1.1: intersection(\#0, \#1) \\
\#\#\# Observation 3.1.1: \#2, which are instances of spaceflight.bipropellant\_rocket\_engine. \\
\#\#\# Thought 3.1.2: \#2 should be the final answer. \\
\# Final answer:\#2.\\
Do you understand it? \\ 
\textbf{Assistant:} Yes, I've understood your instruction and the demonstration.\\
\textbf{User:} Great! The new question is \{question\} \\ \bottomrule
\caption{\texttt{DARA} for ICL Agents}
\label{tab:prompt-dara-icl}
\end{longtable}
\twocolumn

\begin{figure}[tb]
    \includegraphics[width=\columnwidth]{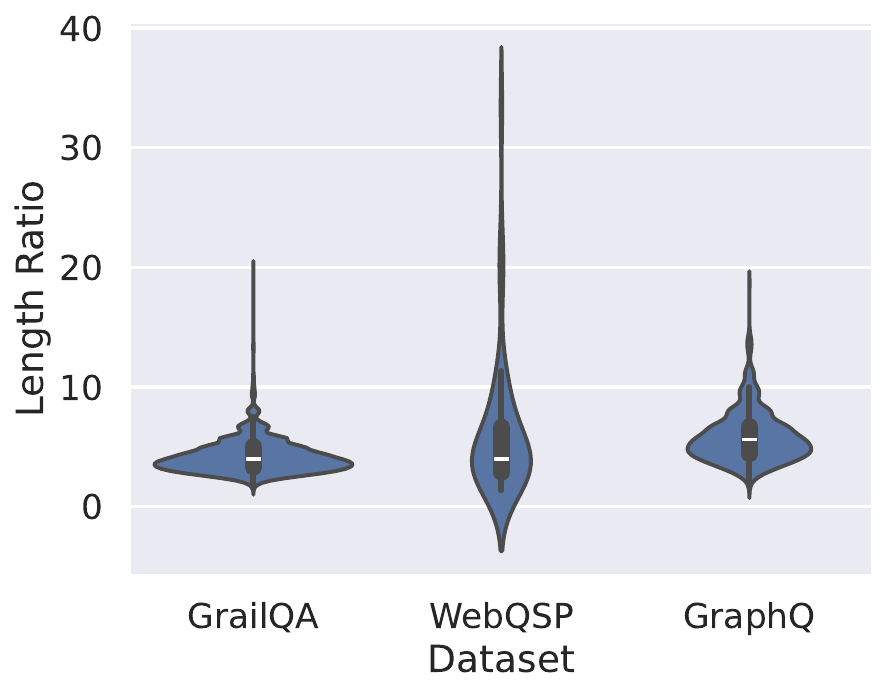}
    \caption{Ratio between tokens of SPARQL and tokens of questions in three test datasets}
    \label{fig:ratio_sparql_q}
\end{figure}
\section{Why is there a performance discrepancy between WebQSP and GrailQA, GraphQ?} \label{app:performance discreapnacy}
One significant factor lies in the different sources of the questions.
Questions of GrailQA and GraphQ are derived from logical forms, which are structured and explicitly aligned with underlying KGs.
Conversely, WebQSP collects questions from real users' query logs on Google search, often containing common sense or complex logical operations.
Intuitively, when the SPARQL is longer but the question is short, it often signifies a requirement for implicit knowledge or complex reasoning. To explore this hypothesis, we simply calculate the ratio between the lengths of SPARQL queries and questions, visualizing it in a violin plot as shown in Figure~\ref{fig:ratio_sparql_q}.
Compared with GrailQA and GraphQ, we observe that WebQSP exhibits more extreme cases in the length ratio, suggesting that it contains more short questions with long SPARQL queries.
Some of them (cf. Table~\ref{tab: length ratio}) include complex logical constraints (e.g. time comparison) that cannot be represented by s-expression.
Besides, Humans can formulate concise and abstract questions based on common sense, which poses a challenge for parsing questions in the absence of such knowledge or an ability to leverage it.
For example, to answer the question \textit{what year did Seattle Seahawks go to the Super Bowl?} The model needs to know Superbowl serves as the final match of the American National Football League. Therefore, it needs to find the season when the Seattle Seahawks were not only the champion but also the runner-up.
Those real questions from humans can be more flexible, and obscurer than synthetic questions derived from logical forms.
In light of these findings, it is evident that advancing the field of knowledge graph reasoning requires the design of more realistic datasets, the creation of more expressive intermediate representations, and the exploration of new model paradigms that can effectively harness common-sense knowledge.
but sometimes such synthetic questions lack clarity and authenticity, which may impede the LLM agent's understanding.
For example, \textit{which industry is ayala land in which real estate and rental and leasing is also in?} is ambiguous and unnatural.

\onecolumn
\begin{longtable}[c]{ll}
\hline
Question &
  what did james k polk do before he was president? \\ \hline
SPARQL &
  \begin{tabular}[c]{@{}l@{}}PREFIX ns: \textless{}http://rdf.freebase.com/ns/\textgreater\\ SELECT DISTINCT ?x\\ WHERE \{\\   \{\\     SELECT ?pFrom \\     WHERE \{\\       ns:m.042f1 ns:government.politician.government\_positions\_held ?y .\\       ?y ns:government.government\_position\_held.office\_position\_or\_title ?x ; \\          ns:government.government\_position\_held.basic\_title ns:m.060c4 ; \# President\\          ns:government.government\_position\_held.from ?pFrom .\\     \}\\   \}\\   ns:m.042f1 ns:government.politician.government\_positions\_held ?y . \# James K. Polk\\   ?y ns:government.government\_position\_held.office\_position\_or\_title ?x ; \\      ns:government.government\_position\_held.from ?from .\\   \\   FILTER(xsd:dateTime(?pFrom) - xsd:dateTime(?from) \textgreater 0)\\ \}\end{tabular} \\ \hline
Question &
  what super bowl did peyton manning win? \\ \hline
SPARQL &
  \begin{tabular}[c]{@{}l@{}}SELECT DISTINCT ?z\\ WHERE \{\\   ns:m.027jv8 ns:sports.pro\_athlete.teams ?y .\\   ?y ns:sports.sports\_team\_roster.team ?x . \#team\\   ?x ns:sports.sports\_team.championships ?z . \# super bowls\\   ?z ns:common.topic.notable\_types ns:m.01xljv1 . \#super bowl\\   ?z ns:sports.sports\_championship\_event.champion ?c .\\   ?z ns:time.event.start\_date ?a .\\   \\   \# Check the time overlap\\   FILTER(NOT EXISTS \{?y ns:sports.sports\_team\_roster.from ?sk0\} || \\              EXISTS \{?y ns:sports.sports\_team\_roster.from ?sk1 . \\              FILTER(xsd:datetime(?sk1) - xsd:datetime(?a) \textless{}= 0) \})\\ \\   FILTER(NOT EXISTS \{?y ns:sports.sports\_team\_roster.to ?sk2\} || \\              EXISTS \{?y ns:sports.sports\_team\_roster.to ?sk3 . \\              FILTER(xsd:datetime(?sk3) - xsd:datetime(?a) \textgreater{}= 0 ) \}) \\ \}\end{tabular} \\ \hline
Question &
  which country in north america is divided into provinces? \\ \hline
SPARQL &
  \begin{tabular}[c]{@{}l@{}}PREFIX ns: \textless{}http://rdf.freebase.com/ns/\textgreater\\ SELECT DISTINCT ?x \\ WHERE \{\\     ns:m.059g4 ns:location.location.contains ?x .  \# North America\\     ?x ns:common.topic.notable\_types ns:m.01mp . \# Country\\     ?x ns:location.location.contains ?y .\\     ?y ns:common.topic.notable\_types ?t . \\ \# All the possible "province" type\\ FILTER ((?t = ns:m.01nm) ||\\ (?t = ns:m.02\_1y\_9) ||\\ (?t = ns:m.02\_3ny\_) ||\\ (?t = ns:m.02\_3phk) ||\\ (?t = ns:m.02\_3r2r) ||\\ (?t = ns:m.02\_3rt3) ||\\ (?t = ns:m.02\_3zf4) ||\\ (?t = ns:m.02\_40h1) ||\\ (?t = ns:m.02\_96lm) ||\\ (?t = ns:m.02yxk5c) ||\\ (?t = ns:m.02zd6yn) ||\\ (?t = ns:m.03z96kq) ||\\ (?t = ns:m.04g7rg9) ||\\ (?t = ns:m.04js0h5) ||\\ (?t = ns:m.065rjpr) ||\\ (?t = ns:m.078\_8dm) ||\\ (?t = ns:m.0hzcb3l) ||\\ (?t = ns:m.0hzcb5p) ||\\ (?t = ns:m.0hzcb69) ||\\ (?t = ns:m.0hzcb7p) ||\\ (?t = ns:m.0hzcd76) ||\\ (?t = ns:m.0hzcd7v) ||\end{tabular} \\ \hline
Question &
  who was president after franklin d. roosevelt? \\ \hline
SPARQL &
  \begin{tabular}[c]{@{}l@{}}PREFIX ns: \textless{}http://rdf.freebase.com/ns/\textgreater\\ SELECT DISTINCT  ?x\\ WHERE \{ \\ \# President of the United States \\   ns:m.060d2 ns:government.government\_office\_or\_title.office\_holders ?y1 . \\  
   \# Franklin D. Roosevelt \\
  ?y1 ns:government.government\_position\_held.office\_holder ns:m.02yy8 ;\\       ns:government.government\_position\_held.to ?to .\\  \# President of the United States \\  ns:m.060d2 ns:government.government\_office\_or\_title.office\_holders ?y2 . \\   ?y2 ns:government.government\_position\_held.office\_holder ?x ;\\       ns:government.government\_position\_held.to ?from .\\   \\   FILTER(xsd:dateTime(?from) - xsd:dateTime(?to) \textgreater 0)\\ \}\\ ORDER BY xsd:dateTime(?from)\\ LIMIT 1\end{tabular} \\ \hline
Question &
  who is meredith gray married to in real life? \\ \hline
SPARQL &
  \begin{tabular}[c]{@{}l@{}}PREFIX ns: \textless{}http://rdf.freebase.com/ns/\textgreater\\ SELECT DISTINCT ?x\\ WHERE \{\\                 ns:m.0268x2f ns:tv.tv\_character.appeared\_in\_tv\_program ?y .\\                 ?y ns:tv.regular\_tv\_appearance.actor ?z .\\                 ?z ns:people.person.spouse\_s ?s .\\                 ?s ns:people.marriage.spouse ?x .\\                 ?s ns:people.marriage.type\_of\_union ns:m.04ztj .\\ Filter (?x != ?z)\\ FILTER(NOT EXISTS \{?s ns:people.marriage.from ?sk0\} || \\            EXISTS \{?s ns:people.marriage.from ?sk1 . \\                    FILTER(xsd:datetime(?sk1) \textless{}= "2015-08-10"\textasciicircum{}\textasciicircum{}xsd:dateTime) \})\\ FILTER(NOT EXISTS \{?s ns:people.marriage.to ?sk2\} || \\            EXISTS \{?s ns:people.marriage.to ?sk3 . \\                    FILTER(xsd:datetime(?sk3) \textgreater{}= "2015-08-10"\textasciicircum{}\textasciicircum{}xsd:dateTime) \})\\ ?s ns:people.marriage.from ?sk4 .\\ \}\\ ORDER BY DESC(xsd:datetime(?sk4))\\ LIMIT 1\end{tabular} \\ \hline
Question &
  what was franklin d roosevelt 's job before president? \\ \hline
SPARQL &
  \begin{tabular}[c]{@{}l@{}}PREFIX ns: \textless{}http://rdf.freebase.com/ns/\textgreater\\ SELECT DISTINCT ?x\\ WHERE \{\\   \{\\     SELECT ?pFrom \\     WHERE \{\\       ns:m.02yy8 ns:government.politician.government\_positions\_held ?y . \# Franklin D. Roosevelt\\       ?y ns:government.government\_position\_held.office\_position\_or\_title ?x ; \\          ns:government.government\_position\_held.basic\_title ns:m.060c4 ; \# President\\          ns:government.government\_position\_held.from ?pFrom .\\     \}\\   \}\\   ns:m.02yy8 ns:government.politician.government\_positions\_held ?y . \# Franklin D. Roosevelt\\   ?y ns:government.government\_position\_held.office\_position\_or\_title ?x ; \\      ns:government.government\_position\_held.from ?from .\\   \\   FILTER(xsd:dateTime(?pFrom) - xsd:dateTime(?from) \textgreater 0\end{tabular} \\ \hline
  \caption{Questions in unseen schemas test data of WebQSP with unseen schemas whose length ratio is larger than 10.}
\label{tab: length ratio}
\end{longtable}
\clearpage
\twocolumn

\section{Error Analysis}\label{app:error anlaysis}
As discussed in section~\ref{sec:error analysis}, we compared \texttt{DARA}(Llama-2-7B) with ICL-based agents (GPT-4) and fine-tuned Agentbench-7B. We show three examples in Table~\ref{tab:cvt} $\sim$ Table~\ref{tab:instruction_follow}.



\begin{table*}[]
\centering
\begin{tabular}{@{}l|l@{}}
\toprule
Question &
  guitar hero iii has what songs? \\ \midrule
ICL Agent (GPT-4) &
  \begin{tabular}[c]{@{}l@{}}
  Thought: I need to find songs that can be played in 'guitar hero 3'. \\ 
  ... \\ 
  Thought: The relation 'cvg.musical\_game.songs' \\ seems to be the most relevant to find the songs in 'guitar hero 3'.\\ Action: get\_neighbors(guitar hero 3, cvg.musical\_game.songs)\\ 
  Observation : variable \#0, which are instances of cvg.musical\_game\_song\_relationship \\
  \textcolor{red}{Thought: The variable \#0 should be the final answer to the question as it represents the songs} \\ \textcolor{red}{that can be played in Guitar Hero 3.}\end{tabular} \\ \midrule
AgentBench-7B & 
    \begin{tabular}[c]{@{}l@{}} Thought: Find the songs that appear in Guitar Hero III.\\
    Action: get\_relations(guitar hero iii)\\
    Observation: ... \\ 
    Thought: use the outgoing relation 'cvg.musical\_game.songs' of Guitar Hero III \\ 
    with the function get\_neighbors to find the songs that appear in Guitar Hero III.\\
    Action: get\_neighbors(guitar hero iii, cvg.musical\_game.songs)\\
    \textcolor{red}{Observation: variable \#0, which are instances of cvg.musical\_game\_song\_relationship} \\
    \textcolor{red}{Final Answer: \#0} \\
    \end{tabular} \\ \midrule
\texttt{DARA} (Llama-2-7B) &
  \begin{tabular}[c]{@{}l@{}}The given question can be decomposed into the following subtasks: \\ \# Task 1: Find the songs that can be played in Guitar Hero 3. \\
  ... \\ 
  \#\#\# Thought 1.1.2: From the explanations, \textcolor{blue}{we can use the outgoing relation} \\ \textcolor{blue}{'cvg.musical\_game.songs' of Guitar Hero 3 with the operator JOIN to reach} \\ \textcolor{blue}{the intermediate (CVT) node first, which contains all related information of songs in the game.} \\
  ...\\ 
  Thought 1.2.2: From the explanations, \\ \textcolor{blue}{we can use the outgoing relation 'cvg.musical\_game\_song\_relationship.song' of} \\ \textcolor{blue}{the CVT node from step 1.1 with JOIN to find the songs that can be played in the game.} \\ ...\end{tabular} \\ \midrule
Explanation & \begin{tabular}[c]{@{}l@{}}
GPT-4 and AgentBench-7B agents prematurely stop at the CVT node while \texttt{DARA} gets \\ the correct answer.\end{tabular}
   \\ \bottomrule
\end{tabular}
\caption{The \textcolor{red}{incorrect reasoning trajectory} is highlighted in red while \textcolor{blue}{the correct one} is in blue.}
\label{tab:cvt}
\end{table*}

\begin{table*}[]
\centering
\begin{tabular}{@{}l|l@{}}
\toprule
Question &
  snk developed how many video game compilations? \\ \midrule
ICL Agent (GPT-4) &
  \begin{tabular}[c]{@{}l@{}}
  
  Thought:  I need to find out the number of video game compilations developed by snk. \\
  To achieve this, I will query the KB to find relations connected to the entity 'snk' and see \\ if any
  of them can help me find the answer. \\
  Thought: From the above relations, I can use cvg.cvg\_developer.games\_developed \\ to find video games developed by snk. \\
  ... \\
  Thought: Now, I need to find out which of these games are compilations. To achieve this, \\I will query the KB to find relations connected to the variable \#0. \\
  \textcolor{red}{Action: get\_relations(\#0)}\\
  ...\\
  \textcolor{red}{Thought: From the above relations, I might use cvg.computer\_game\_compilation} \\ \textcolor{red}{.games\_included to find which of these games are compilations.}
  \end{tabular} \\ \midrule
  AgentBench-7B & \begin{tabular}[c]{@{}l@{}}
       Thought: Find the video game compilations developed by SNK.\\
       Action: get\_relations(snk)\\
       Observation: ... \\
    Thought: Use the outgoing relation 'cvg.cvg\_developer.game\_versions\_developed' of \\
    SNK with the function get\_neighbors to find the video game compilations developed by SNK.\\
    Action: get\_neighbors(snk,cvg.cvg\_developer.game\_versions\_developed)\\
    Observation: variable \#0, which are instances of cvg.game\_version\\
    Thought: Use the function count on the results obtained from \#0 to get the final answer.\\
    Action: count(\#0)
  \end{tabular} \\ \midrule
\texttt{DARA} (Llama-2-7B) &
  \begin{tabular}[c]{@{}l@{}}The given question can be decomposed into the following subtasks: \\ \# Task 1: Find the video game compilations developed by SNK. \\... \\ Thought 1.1.2: From the explanations, use the outgoing relation \\'cvg.cvg\_developer.games\_developed' \\ of SNK with the operator JOIN to find the video game compilations developed by SNK. \\ 
  ... \\
  \#\#\# \textcolor{blue}{Action 1.2.1: get\_classes(s-exp-1.1)} \\
  ...\\
  \#\#\# Thought 1.2.1: From classes in Obs 1.2.1, use the operator AND with the class \\'cvg.computer\_game\_compilation' to filter out the entities that are not video game \\ compilations.\\
  \#\#\# \textcolor{blue}{S-exp-1.2: (AND cvg.computer\_game\_compilation s-exp-1.1)} \end{tabular} \\ \midrule
Explanation & \begin{tabular}[c]{@{}l@{}} GPT-4 and AgentBench-7B agents are incapable of getting class information due to \\ the lack of functions in the action space while \texttt{DARA} calls the correct function.
\end{tabular}
   \\ \bottomrule
\end{tabular}
\caption{The \textcolor{red}{incorrect reasoning trajectory} is highlighted in red while \textcolor{blue}{the correct one} is in blue.}
\label{tab:action space}
\end{table*}

\begin{table*}[]
\centering
\begin{tabular}{@{}l|l@{}}
\toprule
Question &
  what pet diseases have the same cause with canine arthritis? \\ \midrule
ICL Agent (GPT-4) &
  \begin{tabular}[c]{@{}l@{}}
  Thought: I need to first find the cause of canine arthritis. To achieve this, I will query the KB \\
  to find relations connected to the entity 'canine arthritis' and \\ see if any of them can help me find the answer.\\
  Action: get\_relations(canine arthritis)\\
  Observation :...\\
    Thought: From the above relations, I might use \\ base.pethealth.pet\_disease\_or\_medical\_condition.causes \\
    to find the cause of canine arthritis.\\
    ... \\
    \textcolor{blue}{Thought: Now I need to find other pet diseases with the same cause.} \\ To acheive this, I will query the KB to find relations connected to the variable \#0 \\
    Action: get\_neighbors(\#0, base.pethealth.cause.pet\_diseases\_or\_conditions\_caused)\\
    Observation : variable \#1, which are instances of \\
    base.pethealth.pet\_disease\_or\_medical\_condition\\
    \textcolor{blue}{Thought: \#1 is the final answer to the question ...}
    
  \end{tabular} \\ \midrule
  AgentBench-7B & \begin{tabular}[c]{@{}l@{}}
    \textcolor{red}{Thought: Find the pet diseases that have the same cause with canine arthritis.}\\
    Action: get\_relations(canine arthritis)\\
    Observation: ... \\
    Thought: use the outgoing relation \\'base.pethealth.pet\_disease\_or\_medical\_condition.causes'\\ of canine arthritis with the function get\_neighbors to find the pet diseases that have \\ 
    the same cause with canine arthritis.\\
    Action: get\_neighbors(canine arthritis, base.pethealth.\\
    pet\_disease\_or\_medical\_condition.causes)\\
    Observation: variable \#0, which are instances of base.pethealth.cause\\
    \textcolor{red}{Final Answer: \#0} \\
   \end{tabular}\\ \midrule
  
\texttt{DARA} (Llama-2-7B) &
  \begin{tabular}[c]{@{}l@{}}The given question can be decomposed into the following subtasks:\\
  \# Task 1: Find the cause of canine arthritis.\\
  ...\\
  \#\# Step 1.1:\#\#\# Action 1.1.1: get\_relations(m.05gj3zc)\\
  \#\#\# Obs 1.1.1: ...\\
  \#\#\# S-exp-1.1: \\
  (JOIN (R base.pethealth.pet\_disease\_or\_medical\_condition.causes) m.05gj3zc)\\
  \textcolor{blue}{\# Task 2: Find the pet diseases that have the same cause as canine arthritis.}\\
  ...\\
  \#\# S-exp-2: (JOIN (R base.pethealth.cause.pet\_diseases\_or\_conditions\_caused) s-exp-1)\\
  \textcolor{blue}{\# Final s-exp: (JOIN (R base.pethealth.cause.pet\_diseases\_or\_conditions\_caused) s-exp-1)}\\
  \end{tabular}
  \\
 \midrule
Explanation & \begin{tabular}[c]{@{}l@{}} GPT-4 agents and \texttt{DARA} (Llama-2-7B) find the correct answer.\\ AgentBench-7B cannot decompose the question into the correct task.\end{tabular}
   \\ \bottomrule
\end{tabular}
\caption{The \textcolor{red}{incorrect reasoning trajectory} is highlighted in red while \textcolor{blue}{the correct one} is in blue.}
\label{tab:instruction_follow}
\end{table*}

\end{document}